\documentclass[10pt,twocolumn,letterpaper]{article}

\usepackage{iccv}
\usepackage{times}
\usepackage{epsfig}
\usepackage{graphicx}
\usepackage{amsmath}
\usepackage{amssymb}
\usepackage{standalone}

\usepackage{enumitem}
\usepackage{bm}

\usepackage{enumitem}
\usepackage{booktabs}
\usepackage{algpseudocode,algorithm,algorithmicx}

\usepackage[pagebackref=true,breaklinks=true,letterpaper=true,colorlinks,bookmarks=false]{hyperref}

\iccvfinalcopy 

\def\iccvPaperID{3505} 

\ificcvfinal\fi

\title{Evolving Space-Time Neural Architectures for Videos}

\author{
AJ Piergiovanni, Anelia Angelova, Alexander Toshev, Michael S. Ryoo \\
  Google Brain\\
  \texttt{\small\{ajpiergi,anelia,toshev,mryoo\}@google.com} \\
}

\begin{document}

\maketitle
\thispagestyle{empty}

\begin{abstract}

We present a new method for finding video CNN architectures that capture rich \emph{spatio-temporal} information in videos.
Previous work, taking advantage of 3D convolutions, obtained promising results by manually designing video CNN architectures. We here develop a novel evolutionary search algorithm that automatically explores models with different types and combinations 
of layers to jointly learn interactions between spatial and temporal aspects of video representations.
We demonstrate the generality of this algorithm by applying it to two meta-architectures, obtaining new architectures superior to manually designed architectures.
Further, we propose a new component,
the iTGM layer, which more efficiently utilizes its parameters to allow learning of space-time interactions over longer time horizons. The iTGM layer is often preferred by the evolutionary algorithm and allows building cost-efficient networks. 
The proposed approach discovers new and diverse video architectures that were previously unknown. More importantly they are both more accurate and faster than prior models, and outperform the state-of-the-art results on multiple datasets we test, including HMDB, Kinetics, and Moments in Time. 
We will open source the code and models, to encourage  future model development \footnote{Code and models: \href{https://sites.google.com/corp/view/evanet-video}{https://sites.google.com/corp/view/evanet-video}}. 

\end{abstract}

\section{Introduction}


Video understanding tasks, such as video object detection and activity recognition, are important for many societal applications of computer vision including robot perception, smart cities, medical analysis, and more.
Convolutional neural networks (CNNs) have been popular for video understanding, with
many successful prior approaches, including C3D~\cite{tran2014c3d}, I3D~\cite{carreira2017quo}, R(2+1)D~\cite{tran2018closer}, S3D~\cite{xie2017rethinking}, and others~\cite{feichtenhofer2016convolutional,karpathy2014large}. 
These approaches focus on manually designing CNN architectures specialized for videos, for example by extending known 2D architectures such as Inception~\cite{szegedy2016rethinking} and ResNet~\cite{he2016deep} to 3D~\cite{carreira2017quo,tran2018closer}. 
However, designing new, larger or more advanced architectures is a challenging problem, especially as the complexity of video tasks necessitates deeper and wider architectures and more complex sub-modules. 
Furthermore, the existing networks, which are mostly inspired by single-image based tasks, might not sufficiently capture the rich spatio-temporal interactions in video data.

In this work, we present a video architecture evolution approach to harness the rich \emph{spatio-temporal} information present in videos. 
Neural architecture search and evolution have been previously applied for text and image classification~\cite{tan2018mnasnet,zoph2017neural}.
A naive extension of the above approaches to video is infeasible due to the large search space of possible architectures operating on 3D inputs.

To address these challenges we propose a novel evolution algorithm for video architecture search. We introduce a hybrid meta-architecture (`fill-in-the-blanks') model for which the high level connectivity between modules is fixed, but the individual modules can evolve. We apply this successfully to both Inception and ResNet based meta-architectures. We design the search space specifically for video CNN architectures that jointly capture various spatial and temporal interactions in videos. 
We encourage exploration of more \emph{diverse} architectures by applying multiple nontrivial mutations at the earlier stages of evolution while constraining the mutations at the later stages.
This enables discovering multiple, very different but similarly good architectures, allowing us to form a better ensemble by combining them.



\begin{figure*}
    \centering
    \includegraphics[width=0.62\linewidth]{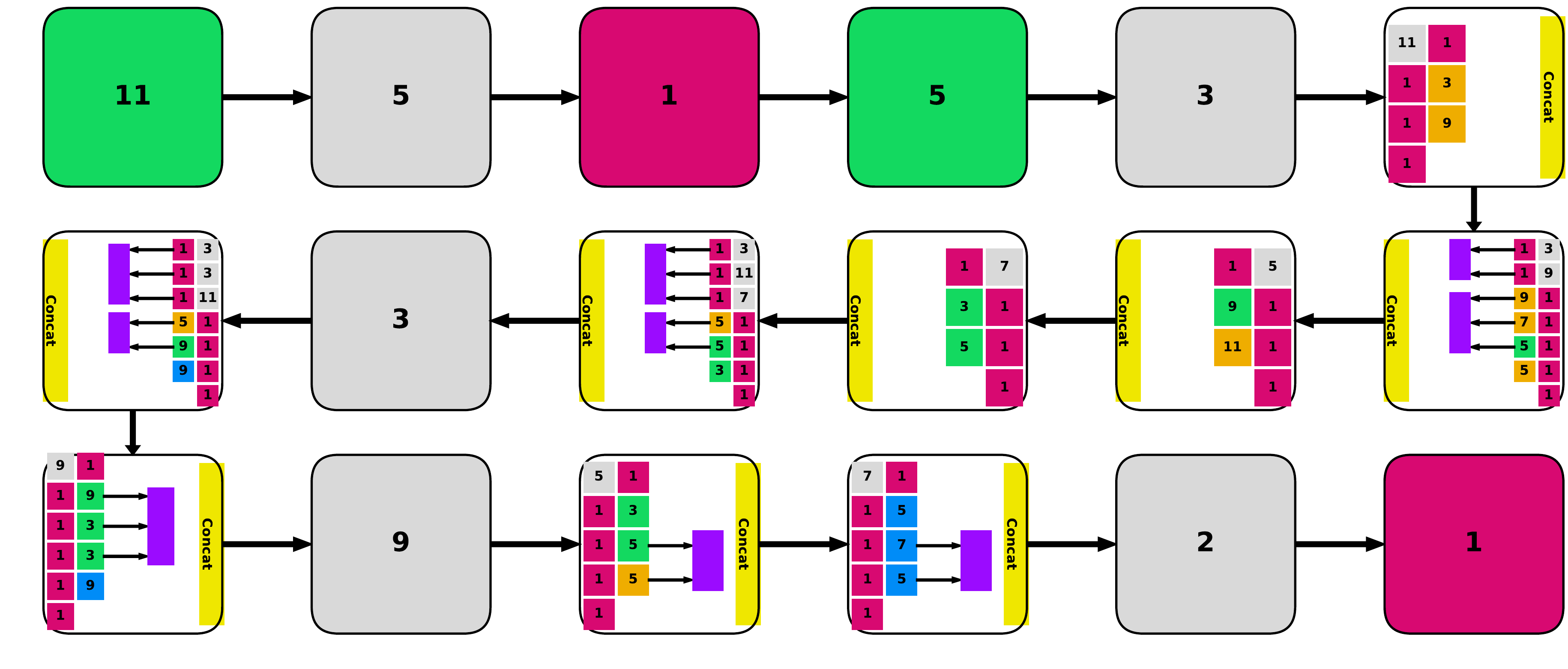} \hspace{8mm} \includegraphics[width=0.2\linewidth]{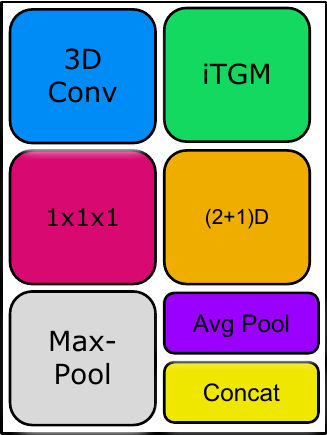} 
    \caption{Example of a video architecture obtained with evolution. Inception-like architecture.  
    The color encodes the type of the layer, as indicated on the right. The numbers indicate the temporal size of the filters in each module. See text for discussion.}
    \label{fig:example-architecture}
\end{figure*}

\begin{figure*}
    \centering
    \includegraphics[width=0.84\linewidth]{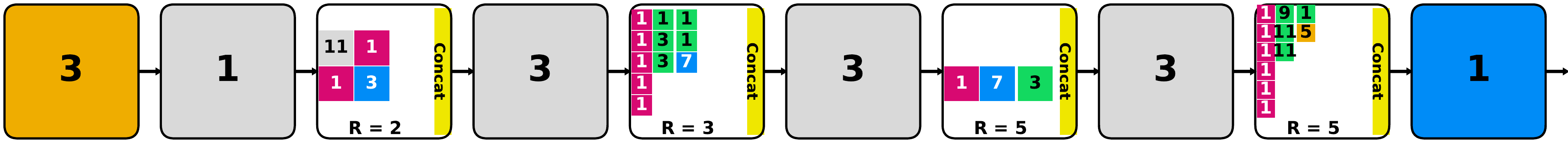}
    \includegraphics[width=0.84\linewidth]{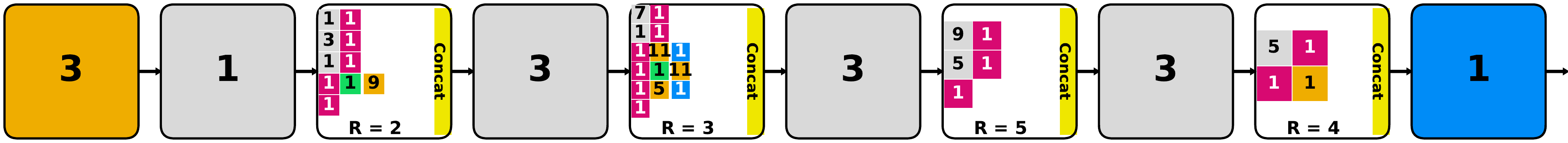}
    \includegraphics[width=0.84\linewidth]{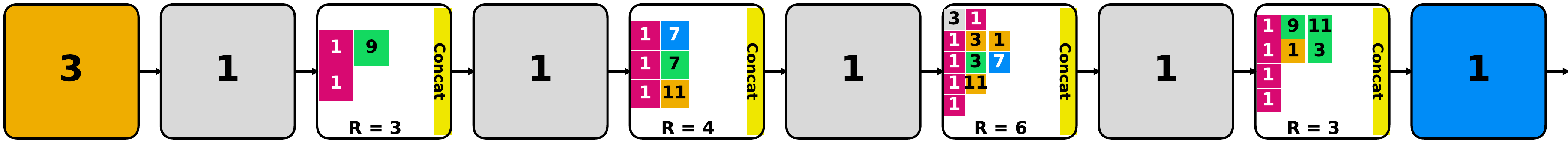}
    \caption{Three different ResNet-like architectures obtained for the Kinetics dataset. Modules are repeated $R$ times. 
    }
    \label{fig:example-architectures-res}
\end{figure*}

Furthermore, to enrich the search space for video inputs, we propose a new key element which is specifically designed to capture space-time features' interactions.
We introduce an \emph{Inflated Temporal Gaussian Mixture} (iTGM) layer as part of the evolution search space. The iTGM is motivated by the original 1D TGM~\cite{tgm}. For our iTGM, we learn 2D spatial filters in addition to the temporal Gaussian mixture values, and inflate the 2D filter temporally to allow learning of joint features in 3D. 
The 2D filter is inflated non-uniformly, by following the weights according to the learned 1-D temporal Gaussian mixture pattern.  This allows to explore space-time interactions more effectively and with much fewer parameters, while at the same time capture longer temporal information in videos.



The proposed algorithm results in novel architectures which comprise interesting sub-modules (see Fig.~\ref{fig:example-architecture} and~\ref{fig:example-architectures-res}). It discovers complex substructures, including modules with multiple parallel space-time conv/pooling layers focusing on different temporal resolutions of video representations. Other findings include: multiple different types of layers combined in the same module e.g., an iTGM layer jointly with (2+1)D convolutions and pooling layers; heterogeneous modules at different levels of the architecture, which is in contrast to previous handcrafted models.
Furthermore, the evolution itself generates a diverse set of accurate models. By ensembling them, recognition accuracy increases beyond other homogeneous-architecture ensembles.

Our approach discovers models which outperform the state-of-the-art on public datasets we tested, including HMDB, Kinetics, and Moments in time. This is done with a generic evolutionary algorithm and no per-data hyperparamter tuning. Furthermore, the best found models are very fast, running at about 100 ms for a single model, and 250ms for an ensemble, both being considerably faster than prior models. 

The main technical contributions of this paper are:
1) We propose a novel evolutionary approach for developing space-time CNN architectures, specifically designed for videos. We design the search space to specifically explore different space-time convolutional layers and their combinations and encourage diversity. 
2) We introduce a new space-time convolutional layer, the Inflated TGM layer, designed to capture longer-term temporal information. 
3) The discovered models achieve state-of-the-art performance on several video datasets and are among the fastest models for videos. We provide new diverse architectures, ensembles and components which can be reused for future work.
To our knowledge this is the first automated neural architecture search algorithm for video understanding.

\vspace{-0.3cm}
\section{Related work}


\paragraph{CNNs for video understanding.}




Approaches considering a video as a space-time volume have been particularly successful \cite{carreira2017quo,hara2017learning,tran2014c3d,tran2017convnet},  with a direct application of 3D CNNs to videos. C3D \cite{tran2014c3d} learned 3x3x3 XYT filters, which was not only applied to action recognition but also to video object recognition. I3D \cite{carreira2017quo} extended the Inception architecture to 3D, obtaining successful results on multiple activity recognition video datasets including Kinetics.
S3D \cite{xie2017rethinking} investigated the usage of 1D and 2D convolutional layers in addition to the 3D layers. R(2+1)D \cite{tran2018closer} used the 2D conv. layers followed by 1D conv. layers while following the ResNet structure. Two-stream CNN design is also widely adopted in action recognition, which takes optical flow inputs in addition to raw RGBs \cite{feichtenhofer2016convolutional,simonyan2014two}. 
There are also works focusing on capturing longer temporal information in continuous videos using pooling \cite{ng2015beyond}, attention \cite{piergiovanni2017learning}, and convolution \cite{karpathy2014large}.
Recurrent neural networks (e.g., LSTMs) are also used to sequentially represent videos \cite{ng2015beyond,yeung2015every}.

\vspace{-10pt}
\paragraph{Neural architecture search.}
Neural network architectures have advanced significantly since the early convolutional neural network concepts of LeCun et al.~\cite{lecun1998gradient} and Krizhevsky et al.~\cite{krizhevsky2012imagenet}: from developing wider modules, e.g., Inception~\cite{szegedy2016rethinking}, or introducing duplicated modules~\cite{lin2013NiN}, residual connections~\cite{he2016deep,xie2017aggregated}, densely connected networks~\cite{huang2017densely,jegou2017tiramisu}, or multi-task architectures: e.g., Faster-RCNN and RetinaNet for detection, and many others~\cite{lin2017focal,luo2018fast,ren2015faster}.
Recently several ground-breaking approaches have been proposed for automated learning/searching of neural network architectures, rather than manually designing them~\cite{real2017large,tan2018mnasnet,zoph2017neural,zoph2017learning}. 
Successful architecture search has been demonstrated for images and text~\cite{zoph2017neural,zoph2017learning}, including object classification. Tran et al.~\cite{tran2018convnet} analyze action recognition experiments with different settings, e.g., input resolution, frame rate, number of frames, network depth, all within the 3D ResNet architecture. 


\vspace{-0.3cm}
\section{Convolutional layers for action recognition}
\label{sec:components}

We first review standard convolutional layers for videos and then introduce the new iTGM layer to learn longer temporal structures with fewer parameters and lower computational cost.
Video CNNs are analogous to standard CNNs, with the difference of an additional temporal dimension in the input and all intermediate feature maps. In more detail, both input and feature maps are represented as 4D tensors XYTC with two spatial dimensions, one temporal and one for the pixel values or features (i.e., channels). Several forms of convolution on such tensors have been explored.



\noindent\textbf{3D convolutional layer} learns a standard 3D convolutional kernel over space and time~\cite{ji20103d}. It applies $C_{out}$ kernels of dimension  $L\times H\times W\times C_{in}$ on a tensor of size $T\times Y\times X\times C_{in}$ to produce a tensor of size $T\times Y\times X\times C_{out}$. This layer has $LHWC_{in}C_{out}$ parameters, which is an order of magnitude larger than CNNs and becomes prohibitive in many cases. Further, expanding 2D kernels to 3D has been explored \cite{mansimov2015initialization}. I3D expanded kernels by stacking the 2D kernels $L$ times, results in state-of-the-art performance~\cite{carreira2017quo}.


\noindent\textbf{(2+1)D convolutional layer} decomposes a 3D kernel into a composition of a 2D spatial kernel followed by a 1D temporal kernel \cite{tran2018closer,xie2017rethinking}. It has $HWC_{in}C_{out} + LC_{out}C_{out}$ parameters, and as such is more efficient than 3D convolution. However, it still depends on the time dimension $L$ which limits the temporal size of the filter.


\begin{figure}
    \centering
    \includegraphics[width=\linewidth]{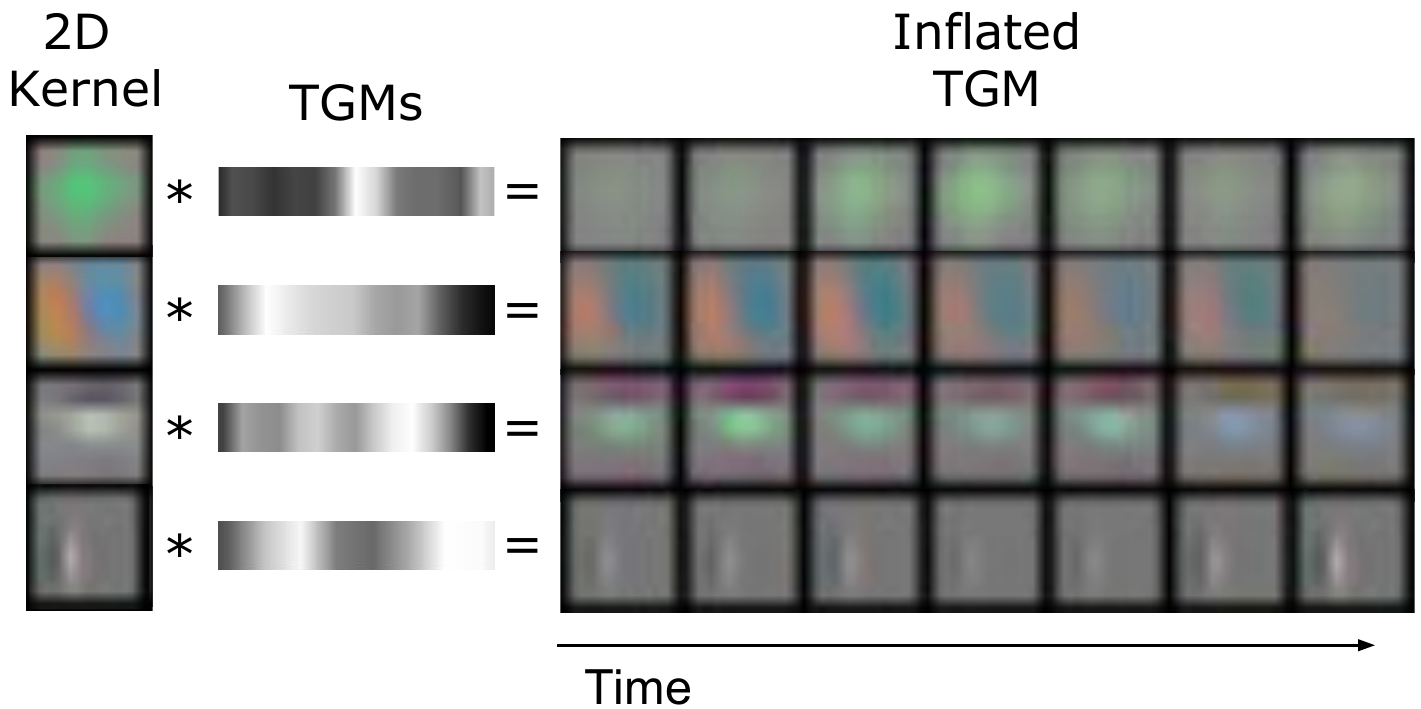}
    \caption{The iTGM layer. Example of inflated TGM kernels.}
    \label{fig:tgm-inflation-imagenet}
\end{figure}

\subsection{3D Inflated TGM layer}
The recently introduced Temporal Gaussian Mixture layer (TGM) ~\cite{tgm} is a specialized 1D convolutional layer designed to overcome the limitations of standard 1D convolutional layers. In contrast to the standard 1D temporal convolutional layer, which was often used in video CNNs such as  R(2+1)D, a TGM layer represents its filter as a mixture of 1D Gaussians. This makes the number of its learnable parameters independent of the temporal filter size; with a TGM layer, one does not have to handle all kernel weights but only the Gaussian mixture parameters.



In this work, we employ the above idea to define a 3D space-time kernel directly, named \emph{Inflated Temporal Gaussian Mixture} layer (iTGM). We `inflate' the 2D spatial kernels to 3D by representing 3D kernel as a product of two kernels:
\begin{equation*}
    S \star K
\end{equation*}
where S is the `inflated' 2D convolution and K is a temporal 1D kernel defined using a mixture of Gaussians~(see Fig.~\ref{fig:tgm-inflation-imagenet}).


The Gaussian mixture kernel $K$ is defined as follows. Denote by $\mu_m$ and width $\sigma_m$ the center and width of $M$ Gaussians, $m\in \{0,\ldots,M\}$. Further, denote by $a_{im}$, $i\in \{0,\ldots,C_{out}\}$ soft-attention mixing-weights.
The temporal Gaussian kernels read:
\begin{equation}
    \hat{K}_{ml} = \frac{1}{Z}\exp\left(-\frac{(l-\mu_m)^2}{2\sigma_m^2}\right)
\end{equation}
where $Z$ is a normalization: $\sum_{l=0}^L \hat{K}_{ml} = 1$. 
Then, the a mixture of the above Gaussian kernels is:
\begin{equation}
\begin{split}
    K_{il} = \frac{\exp{(a_{im})}}{\sum_j \exp{(a_{ij})}}\hat{K}_{ml}.
\end{split}
\end{equation}

This results in $K$ being a $C_{out}\times L$ kernel; i.~e.,~a temporal kernel with $C_{out}$ output channels. We apply this kernel on the output of the spatial kernel. Thus, we obtain a $L\times H\times W\times C_{in}\times C_{out}$ kernel, using only $HWC_{in}C_{out} + 2M + MC_{out}$ parameters.

In practice, $\mu$ is constrained to be in $[0, L)$,  $\mu = (1/2)(L-1)\tanh{(\hat{\mu}})+1$. and $\sigma$ is positive, $\sigma^2 = \exp{(\hat{\sigma})}$. Further, $M$ is a hyperparameter, typically smaller than $L$.


The parameters of the iTGM layer -- spatial kernel parameters, $\mu_m$, $\sigma_m$, and $a_{im}$ -- are all differentiable, and are learned from data for the specified task. The above layer behaves exactly like the standard 3D XYT convolution. Note that this layer learns fewer parameters than both 3D and (2+1)D convolution, and can learn temporally longer kernels as the number of parameters is independent of the length, $L$.
Examples of inflated TGMs are shown in Fig.~\ref{fig:tgm-inflation-imagenet}.

\begin{figure*}
    \centering
    \includegraphics[width=\linewidth]{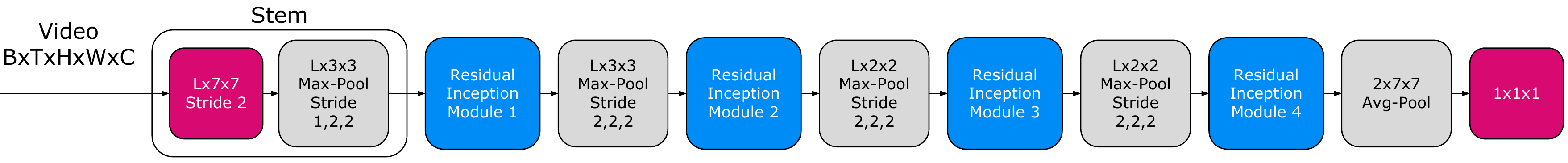}
    \caption{Our ResNet-like `fill-in-the-blanks' meta-architecture: each heterogeneous module is repeated $R$ times, based on the evolution.}
    \label{fig:inception-network}
\end{figure*}



\section{Neural architecture evolution for videos}


We design our neural architecture search specifically for videos, and propose the following: 
\begin{itemize}[nosep,leftmargin=*]
    \item Use of `fill-in-the-blanks' meta-architectures to limit the search space and generate both trainable and high-performing architectures.
    \item Search among combinations of six different types of space-time convolution/pooling layer concatenations where their temporal duration can vary in large ranges. 
    \item We specially design mutation operations to more effectively explore the large space of possible architectures.
    \item We propose an evolutionary sampling strategy which encourages more diverse architectures early in the search.
\end{itemize}

Neural architecture evolution finds better-performing architectures by iteratively modifying a pool of 
architectures. 
Starting from a set of random architectures, it mutates them over multiple rounds, while only retaining the better performing ones. 
Recent studies \cite{real2018regularized} show that evolutionary algorithms can find good image architectures from a smaller number of samples, as opposed to model search algorithms using reinforcement learning~\cite{zoph2017neural}. 
This makes evolution more suitable for video architecture search, as video CNNs are expensive to train. 
Further, it allows for mutating architectures by selecting and combining various space-time layers which more effectively process inputs with much larger dimensionality.
The evolution also enables obtaining multiple different architectures instead of a single architecture which we use to build a powerful ensemble.





\subsection{Search space and base architecture}
\label{subsec:space}

\begin{figure}
    \centering
    \includegraphics[width=0.5\linewidth]{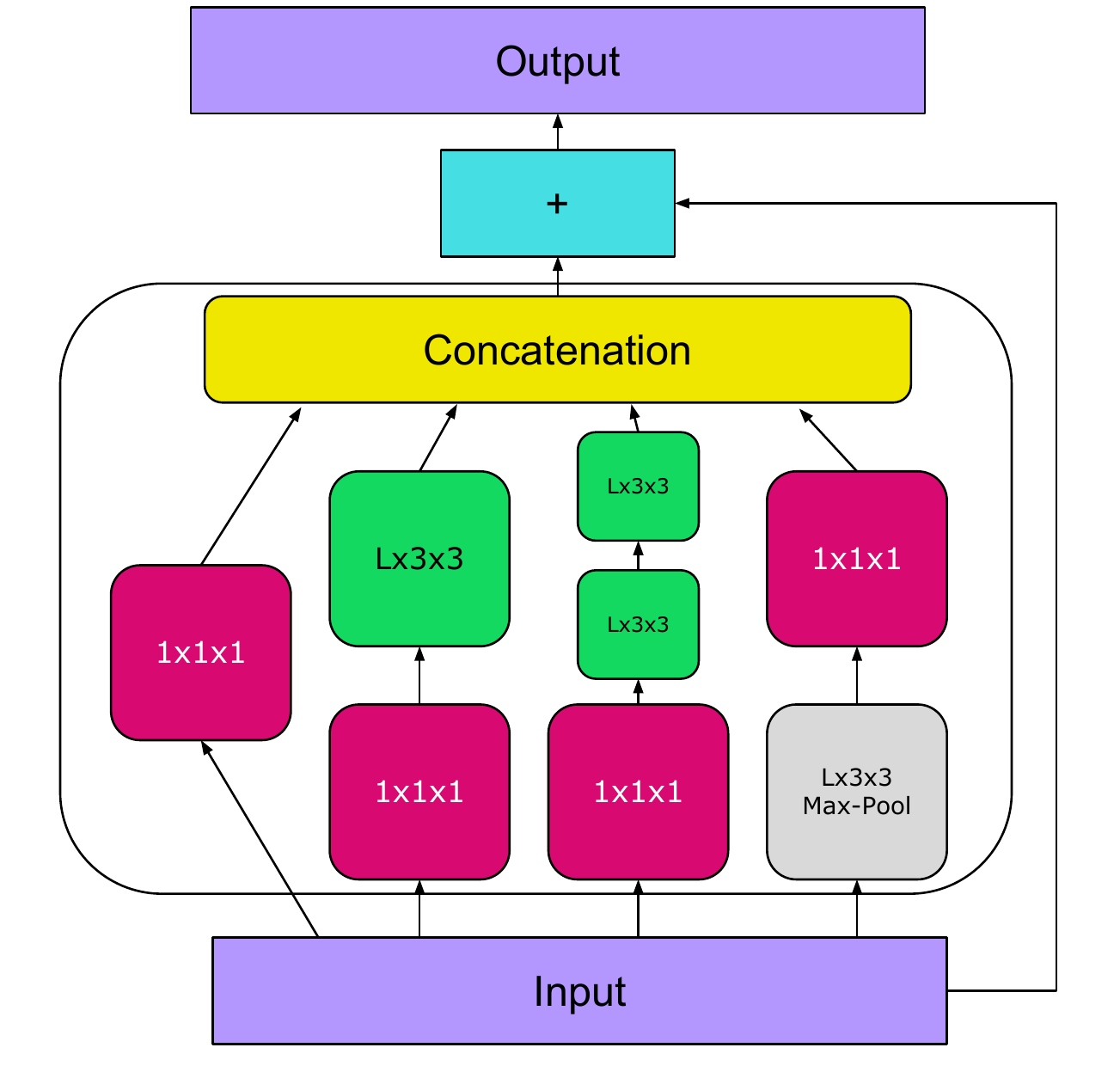}
    \caption{A example structure of the a residual Inception module with 4 layer streams. There could be 1-6 parallel streams (with 4 types) and a residual connection from input to output.}
    \label{fig:inception-module}
\end{figure}

We evolve our architectures to have heterogeneous modules, motivated by the recent observations that video CNN architectures may need differently sized temporal filters at different layers, e.g., bottom-heavy vs.~top-heavy \cite{xie2017rethinking}. 
In order to keep the entire search space manageable while evolving modules heterogeneously, we use a meta-architecture where internal sub-modules are allowed to evolve without constraints but the high level architecture has a fixed number of total modules. We used both an Inception-like and ResNet-like meta-architecture. 
The Inception meta-architecture follows the popular Inception architecture, with five layers forming the `stem' followed by Inception modules whose structure is evolved.
The ResNet-like meta-architecture is illustrated in Figure \ref{fig:inception-network}. This meta-architecture 
is composed of two fixed convolutional layers (i.e., the `stem') followed by four residual Inception modules interspersed with max-pooling layers. Each residual Inception module can be repeated $R$ times and has a residual connection.  
Figure \ref{fig:inception-module} shows an example module.

Each module can have multiple parallel convolutional or pooling layers and its specific form is chosen through evolution. 
We constrain the complexity of the connections between the layers within a module while making the evolution explore temporal aspects of the modules. More specifically, we make each module have 1-6 parallel `streams' with four different stream types: (i) one 1x1x1 conv., (ii) one space-time conv.~layer after one 1x1x1 layer, (iii) two space-time conv.~layers after one 1x1x1, and (iv) a space-time pooling followed by one 1x1x1. Figure \ref{fig:inception-module} shows the four types. 
The architecture evolution focuses on modifying each module: selecting layer types and its parameters, selecting the number of parallel layers, and for the residual ones, how many times should each module be repeated.


The convolutional layers have \{1, 3, 5, 7, 9, 11\} as the set of possible temporal kernel sizes. As a result, the architecture search space size is $O((3 \times 6 + 1)^{5 + B\times N} + (6+1)^{D\times N})$ where $B$ and $D$ are the maximum of number of space-time conv and pooling layers we allow in each module, and $N=4$ or $9$ is the number modules in the meta-architecture. There are $2$ or $5$ individual layers (often also called a `stem') before the modules. Each space-time conv. layer has $3 \times 6$ possible options and each space-time pooling has $6$ options.  Also, there is the option to add/omit the layer, making the total number of choices $3 \times 6 + 1$ and $6+1$. For the ResNet-like models, we allow modules to be repeated up to 6 times.
We fix the spatial size of the kernels to be $3\times 3$.
Although the search space is very big, the idea is that an exhaustive search is not necessary and it is possible to find good local optima by evolving from various initial samples (i.e., architectures).

\subsection{Evolutionary algorithm}

Algorithm \ref{alg:search} summarizes the architecture search. In a standard genetic algorithm setting, we maintain a population of size $P$, where each individual in the population is a particular architecture. Initial architectures are obtained by randomly sampling from our large search space, encouraging diversity and exploration.
At each round of the evolution, the algorithm randomly selects $S$ number of samples from the entire population and compares their recognition performance. The architecture with the highest fitness (i.e., validation accuracy) becomes the `parent', and mutation operators are applied to the selected parent to generate a new `child' architecture to be added to the population. Whenever a new architecture is added, it is trained with the training set for a number of iterations, and is evaluated with a separate validation set (different from the actual test and validation sets) to measure the recognition accuracy. This performance becomes the `fitness' of the architecture. 
Having $S$ where $1 < S \leq P$ controls the randomness in of the parent selection. It avoids the algorithm repeatedly selecting the same parent, which might already be at a local maximum. 

\begin{algorithm}  
  \caption{Evolutionary search algorithm
    \label{alg:search}}  
  \begin{algorithmic}  
    \Function{Search}{}
    \State Randomly initialize the population, $P$
    \State Evaluate each individual in $P$
    \For{$i <$ number of evolutionary rounds}
       \State $S = $ random sample of 25 individuals
       \State $parent = $ the most fit individual in $S$
       \State $child = parent$
       \For{$\max(\lceil d - \frac{i}{r} \rceil, 1)$} \State $child = mutate(child)$ \EndFor
       \State evaluate $child$ and add to population
       \State remove least fit individual from population
    \EndFor
    \EndFunction
  \end{algorithmic}  
\end{algorithm}


\vspace{-15pt}
\paragraph{Mutations.}

\begin{figure}
    \centering
    \includegraphics[width=0.90\linewidth]{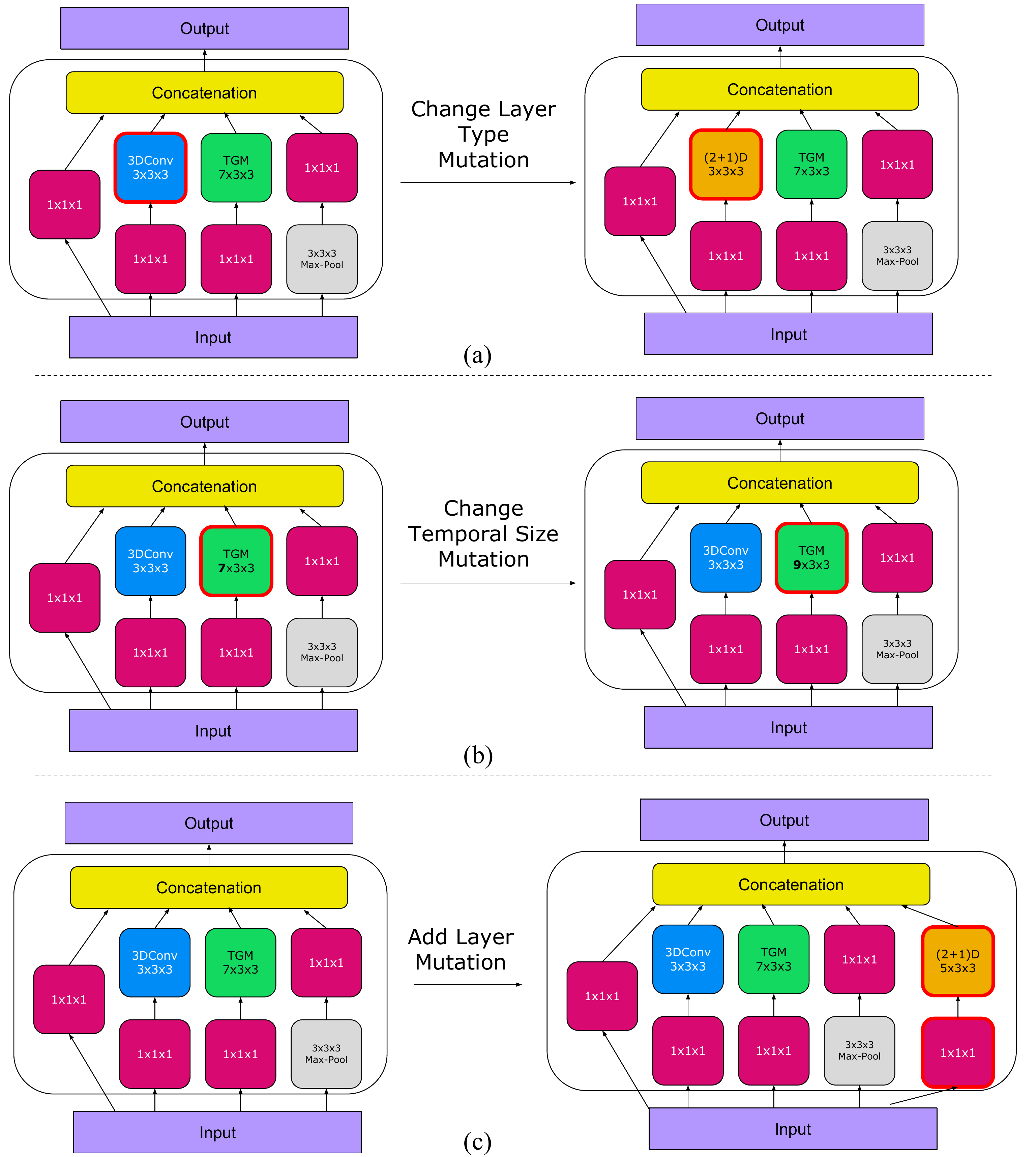}
    \caption{Example mutations applied to a module, including (a) layer type change, (b) filter length change, and (c) layer addition.}
    \label{fig:mutations}
\end{figure}

The mutation operators modify the parent architecture to generate a new child architecture.
In order to explore the architecture search space we describe in Section \ref{subsec:space} efficiently, we consider the following 4 mutation operators:
(i) Select a space-time conv. layer within the parent architecture, and change its `type'.
(ii) Select a space-time conv. layer or a pooling layer, and change its temporal size (i.e., $L$) .
(iii) Select a module from the parent architecture, and add/remove a parallel layer stream.
We constrain the number of parallel layer streams to be 1-6.
We additionally constrain each module to have a fixed number of output filters which are evenly divided between the parallel layers.
(iv) Select a module and change the number of times it is repeated.
Figure \ref{fig:mutations} illustrates examples of our mutation operators applied to layers of a module.

\vspace{-15pt}
\paragraph{Diversity.}
Importantly, we design the mutation in our algorithm to happen by applying multiple randomly chosen mutation operators. In order to encourage more diverse architectures, we develop the strategy of applying many mutation operators in the early stage of the evolution while reducing the amount of mutations in the later stages, which is analogous to controlling the learning rate in a CNN model learning. As described in Algorithm \ref{alg:search}, we apply $\max(d - \frac{i}{r}, 1)$ number of mutation operators where $d$ is the maximum number of operators we want to apply in the beginning, and $r$ controls how quickly we want to decrease their numbers linearly.
Once a child architecture is added to the population, in order to maintain the size of the population to $P$, the evolutionary algorithm selects an individual to discard from the pool. We tried different removal criteria including the lowest fitness and the oldest (i.e., \cite{real2018regularized}), which did not make much difference in our case.

\vspace{-15pt}
\paragraph{Ensemble.}

We obtain a number of top performing architectures after the evolutionary search is completed, thanks to our evolutionary algorithm promoting populations with diverse individual architectures. 
Thus, we are able to construct a strong ensemble from the diverse models by averaging the outputs of their softmax layers: 
$F^*(x) = \sum_{i} F_i(x)$ where $x$ is the input video and $F_i$ are the top models.
In the experiments, we found our approach obtains very diverse, top performing architectures. Ensembling further improves the overall recognition.
We named our final ensemble network as EvaNet (Evolved Video Architecture).


\section{Experiments}

Although our evolutionary architecture search is applicable to various different video understanding tasks, here we focus on human activity recognition. 
The video CNN architectures are evolved using public datasets. Fitness of the architectures during evolution is measured on a subset of the training data. In all experiments, the evolutionary algorithm has no access to the test set during training and evolution. In more detail, we use following datasets:


\noindent\textbf{HMDB~\cite{hmdb}} is a dataset of human motion videos collected from a variety of sources. It is a common datasets for video classification and has $\sim$7K videos of 51 action classes.\newline
\textbf{Kinetics~\cite{kay2017kinetics}} is a large challenging video dataset with 225,946 training and 18,584 validation videos.
We use the currently available version (Kinetics-400 dataset), which has about 25k fewer training videos than original Kinetics dataset (i.e., missing about 10\% of train/val/test data). This makes the dataset more difficult to train, and not comparable to the previous version. \newline
\textbf{Charades~\cite{sigurdsson2016hollywood}} is an activity recognition dataset with $\sim$10K videos, whose durations are 30 seconds on average. We chose Charades to particularly confirm whether our architecture evolution finds structures different from those found with shorter videos like Kinetics. We use the standard classification evaluation protocol. \newline
\textbf{Moments in Time~\cite{monfortmoments}} is a large-scale dataset for understanding of actions and events in videos (339 classes, 802,264 training, 33,900 validation videos).

\subsection{Experimental setup}

Architecture evolution is done in parallel on smaller input size and fewer number of iterations. Details can be found in the appendix. 
We perform evolution for 2000 rounds: generating, mutating, training/evaluating, and discarding 2000 CNN architectures.
Note that $\sim$300 rounds were often sufficient to find good architectures (Figure \ref{fig:evolution-vs-random}).
Once the architecture evolution is complete and the top performing models are found, they are trained on full inputs. 

\noindent\textbf{Baselines.}
We compare our results to state-of-the-art activity recognition methods. 
We train (1) the original I3D~\cite{carreira2017quo} with standard 3D conv.~layers. We also train an Inception model with: (2) 3D conv.~layers with $L=3$, (3) (2+1)D conv.~layers, and (4) the proposed iTGM layers. 
The difference between (1) and (2) is that (1) uses $L=7$ in the first 3D conv.~layer and $L=3$ in all the other 3D layers (a handcrafted design), while (2) uses $L=3$ in all its layers. 


\subsection{Results}
\label{subsec:results}
Next, we report the results of the proposed method and compare with baselines and prior work. This is not only done in terms of recognition accuracy but also in terms of computational efficiency. As shown in Table~\ref{tab:runtime}, our individual models are 4x faster and the ensemble (EvaNet) is 1.6x faster than standard methods like ResNet-50. 
Both of our meta-architectures perform similarly. Below, we report results of the ResNet-like architecture (see suppl.~material for further results).

\begin{table}
  \caption{HMDB split 1 comparison to baselines, with and without Kinetics pre-training. The models were all initialized with ImageNet weights. 
  }
  \label{tab:hmdb-simple-mutations}
  \centering
  \setlength{\tabcolsep}{2pt}
  \begin{tabular}{lccc||ccc}
    \toprule
            &   \multicolumn{3}{c}{HMDB} & \multicolumn{3}{c}{HMDB(pre-train)} \\
            &  RGB & Flow & RGB+F & RGB & Flow & RGB+F \\
    \midrule
    \multicolumn{7}{l}{Baselines} \\
    I3D      & 49.5  & 61.9 & 66.4  & 74.8 & 77.1 & 80.1 \\
    3D Conv  & 47.4  & 60.5 & 65.9  & 74.3 & 76.8 & 79.9 \\
    (2+1)D Conv & 27.8 & 56.4 & 51.8 & 74.4 & 76.5 & 79.9 \\
    iTGM Conv  & 56.5 & 62.5 & 68.2  & 74.6 & 76.7 & 79.9\\
    \hline
    3D-Ensemble & & & 67.6 & & & 80.4\\
    iTGM-Ensemble & & & 69.5 & & & 80.6\\
    
    \midrule
    \multicolumn{7}{l}{Top individual models from evolution}\\
    Top 1 & 60.7 & 63.2 & 70.3 & 74.4 & 78.7 & \textbf{81.4}\\
    Top 2 & 63.4 & 62.5 & \textbf{71.2} & 75.8 &	78.4 & 80.6\\
    Top 3 & 60.5 & 63.1 & 70.5 & 75.4 & 78.9 & 79.7 \\
    \midrule
    EvaNet &  &  & \textbf{72.8} &  &  & \textbf{82.7}\\
    \bottomrule
  \end{tabular}
\end{table}

\begin{table}
  \caption{HMDB performances averaged over the 3 splits.}
  \label{tab:hmdb-sota}
  \centering
  \setlength{\tabcolsep}{10pt}
  \begin{tabular}{lc}
    \toprule
    Two-stream~\cite{simonyan2014two}   & 59.4\\
    Two-stream+IDT~\cite{feichtenhofer2016convolutional}   & 69.2\\
    R(2+1)D~\cite{tran2018closer}   & 78.7\\
    Two-stream I3D~\cite{carreira2017quo}  & 80.9\\
    PoTion~\cite{choutas2018potion}  & 80.9\\
    Dicrim. Pooling~\cite{wang2018discpool} & 81.3\\ 
    DSP~\cite{wang2018dsp} & 81.5\\
    Top model (Individual, ours) & 81.3 \\
    \midrule
    3D-Ensemble    & 79.9\\
    iTGM-Ensemble  & 80.1\\
    EvaNet (Ensemble, ours) & \textbf{82.3}\\
    \bottomrule
  \end{tabular}
\end{table}

\textbf{HMDB:}
Table~\ref{tab:hmdb-simple-mutations} shows the accuracy of the evolved CNNs compared to the baseline architectures, where the evaluation is done on `split 1'.
We see improved accuracy of our individual models as well as ensembles. We also confirm that the EvaNet ensemble is superior to the ensembles obtained by combining other architectures (e.g., 3D ResNet).
Table~\ref{tab:hmdb-sota} compares our performance with the previous state-of-the-arts on all three splits following the standard protocols. 
As seen, our EvaNet models have strong performances outperfoming the state-of-the-art.


\textbf{Kinetics:} 
Table~\ref{tab:kinetics-sota} shows the classification accuracy of our algorithm on Kinetics-400, 
and compares with baselines, other ensembles, and the state-of-the-art.
The architecture evolution finds better performing models than any prior model.  
Further the ensemble of 3 models (EvaNet) improves the performance and outperforms other ensembles, including and ensemble of diverse, standard architectures.

\begin{table}
\caption{Performances on Kinetics-400 Nov. 2018 version. Note that this set is $\sim$10\% smaller (in training/validation set size) than the initial version of Kinetics-400. We report the numbers based on models trained on this newest version. Baselines are shown on top, followed by the state-of-the-arts, and then our methods.}
\label{tab:kinetics-sota}
\centering
\setlength{\tabcolsep}{3pt}
\begin{tabular}{lc}
\toprule
Method & Accuracy \\
 
\midrule
3D Conv   & 72.6 \\
    (2+1)D Conv  & 74.3 \\
    iTGM Conv   & 74.4 \\
ResNet-50 (2+1)D & 72.1  \\
ResNet-101 (2+1)D & 72.8   \\
\midrule
    
    3D-Ensemble    &74.6 \\
    iTGM-Ensemble   & 74.7 \\
    Diverse Ensemble (3D, (2+1)D,  iTGM)  & 75.3 \\
    \midrule
Two-stream I3D \cite{carreira2017quo}  & 72.6   \\ 
Two-stream S3D-G \cite{xie2017rethinking}  & 76.2 \\
ResNet-50 + Non-local \cite{wang2017nonlocal}  & 73.5  \\
Arch. Ensemble (I3D, ResNet-50, ResNet-101) & 75.4 \\
\midrule
    Top 1 (Individual, ours) & \textbf{76.4} \\
    Top 2 (Individual, ours) & 75.5 \\
    Top 3 (Individual, ours) & 75.7 \\
\midrule
    Random Ensemble  & 72.6 \\
\midrule

\midrule
EvaNet (Ensemble, ours) & \textbf{77.2}  \\
\bottomrule
\end{tabular}
\end{table}

\textbf{Charades:}
We also test our approach on the popular Charades dataset. Table~\ref{tab:charades-sota} compares against the previously reported results (we use Kinetics pre-training as in~\cite{wang2017nonlocal}). As shown, we outperform the state-of-the-art and establish a new one with our EvaNet. Our CNNs only use RGB input (i.e., one-stream) in this experiment.

\begin{table}
  \caption{Charades classification results against state-of-the-arts.}
  \label{tab:charades-sota}
  \centering
  \begin{tabular}{lc}
    \toprule
     & mAP \\
    \midrule
    Two-Stream \cite{sigurdsson2016asynchronous} & 18.6\\
    Two-Stream + LSTM \cite{sigurdsson2016asynchronous} & 17.8\\
    Async-TF \cite{sigurdsson2016asynchronous} & 22.4\\
    TRN \cite{zhou2017temporal} & 25.2\\
    Dicrim. Pooling \cite{wang2018discpool} & 26.7\\ 
    Non-local NN \cite{wang2017nonlocal} & 37.5\\ 
    \midrule
    3D-Ensemble (baseline)   & 35.2\\
    iTGM-Ensemble (baseline) & 35.7\\
    \midrule
    Top 1 (Individual, ours) & \textbf{37.3 } \\
    Top 2 (Individual, ours) & 36.8 \\
    Top 3 (Individual, ours) & 36.6 \\

    \midrule
    EvaNet (Ensemble, ours) & \textbf{38.1}\\

    \bottomrule
  \end{tabular}
\end{table}

\textbf{Transfer learned architectures - Moments in Time:} We evaluate the models evolved on Kinetics by training it on another dataset: Moments in Time \cite{monfortmoments}. Table \ref{tab:moments} shows the results, where we see that the models outperform prior methods and baselines. This is particularly appealing as the evolution is done on another dataset and successfully transfers to a new dataset.

\begin{table}
\caption{Moments in time. We show that models evolved on Kinetics transfer to similar datasets.}
\label{tab:moments}
\centering
\setlength{\tabcolsep}{2pt}
\begin{tabular}{lc}
\toprule
Method & Accuracy \\
 
\midrule
    I3D \cite{monfortmoments}   & 29.5 \\
    ResNet-50  & 30.5 \\
    ResNet-50 + NL \cite{wang2017nonlocal}   & 30.7\\
    Arch. Ensemble (I3D, ResNet-50, ResNet-101) & 30.9 \\
   \midrule
    Top 1 (Individual, ours) & \textbf{30.5} \\
   \midrule
EvaNet (Ensemble, ours) & \textbf{31.8}  \\
\bottomrule
\end{tabular}
\end{table}

\begin{table}
\caption{Test accuracy across datasets for a model evolved on a single dataset.}
\label{tab:search-transfer}
\small
\centering
\begin{tabular}{l|cccc}
\toprule
Method & Kinetics & Charades & HMDB & MiT \\
\midrule
Evolved on Kinetics & \textbf{77.2}  & 37.8 & \textbf{82.3} & \textbf{31.8} \\
Evolved on Charades & 76.5 & \textbf{38.1} & 81.8 & 31.1 \\
Evolved on HMDB  & 77.0 & 37.5 & \textbf{82.3} & 31.6 \\
Best without evolution & 76.2 & 37.5 & 81.5 & 30.7 \\
\bottomrule
\end{tabular}
\end{table}

\textbf{Ensembling and runtime.}
One key benefit of evolving model architectures is that the resulting models are naturally diverse, as they are evolved from very different initial random models. 
As shown in Table~\ref{tab:kinetics-sota}, we compared  with an ensemble of three different baselines (3D Conv + (2+1)D + iTGM) and with an ensemble of  different architectures (e.g., I3D + ResNet-50 + ResNet-101). Both are outperformed by EvaNet, although the base models are individually strong.

Furthermore, our evolved models are very efficient performing inference on a video in $\sim$100 ms (Table~\ref{tab:runtime}).
Note that even an ensemble is faster, 258 ms, than previous individual models which makes the proposed approach very suitable for practical use with higher accuracy and faster runtimes. This gain in runtime is due to the use of parallel shallower layers and the use of iTGM layers, which is by itself faster than prior layers (274ms vs 337ms).

\begin{table}
\caption{Runtime measured on a V100 GPU. Accuracy numbers on Kinetics-400 are added for context. These numbers are evaluation time for 1 128 frame clip at 224x224.}
\label{tab:runtime}
\centering
\begin{tabular}{lcc}
\toprule
Method  & Accuracy & Runtime\\
\hline
I3D & 72.6 & 337ms\\
S3D & 75.2 & 439ms\\
ResNet-50 & 71.9 & 526ms \\
ResNet-50 + Non-local & 73.5  & 572ms\\
\hline
I3D iTGM (ours) & 74.4 & 274ms \\
\hline
Individual learned model (ours) & 75.5  & \textbf{108ms}\\
EvaNet (Ensemble, ours) & 77.2 & \textbf{258ms}\\
\hline
\end{tabular}
\end{table}

\noindent\textbf{Architecture findings.}
Figures \ref{fig:example-architecture} and \ref{fig:example-architectures-res} show examples of the architectures found. 
Interesting substructures discovered include: (1) modules combining multiple space-time pooling layers with different temporal intervals and (2) modules heavily relying on Inflated TGM or (2+1)D conv. layers instead of standard 3D conv. layers. Such modules were commonly observed at most of the locations in the architectures, while being very diverse and heterogeneous. 

Video CNN architectures may evolve differently depending on the datasets. This is as expected, and we were able to explicitly confirm this. The architectures have many more layers with longer space-time filters (e.g., 9 or 11) when evolved for Charades, while they only had a small number of them when evolved for HMDB or Kinetics. An average activity duration in Charades videos are around 12 seconds, while HMDB and Kinetics videos are on the average of 3 to 5 seconds. Different architectures are needed for different datasets/tasks, and we are providing an evolutionary approach to automate the architecture design.

Table~\ref{tab:v1_v2} further shows that both Inception-like and ResNet-like meta-architectures are successful, and a combination of them is even more successful.

\begin{table}
\caption{Comparison between models from different hybrid meta-architectures. Kinetics dataset.}
\label{tab:v1_v2}
\centering
\begin{tabular}{lc}
\toprule
Method & Accuracy \\
\midrule
EvaNet Inception (Ensemble, ours) & 76.8  \\
EvaNet ResNet (Ensemble, ours) & 77.2  \\
EvaNet Combined (Ensemble, ours) & \textbf{77.4}  \\
\bottomrule
\end{tabular}
\end{table}

\subsection{Ablation Studies}

\begin{table}
\footnotesize
  \caption{Statistics of the top models. iTGM layers are most common and have longest temporal duration. Kinetics dataset.}
  \label{tab:model-stats}
  \centering
  \begin{tabular}{lccc|cccc}
    \toprule
    & \multicolumn{3}{c|}{Number of Layers} & \multicolumn{4}{c}{Ave.~Temporal Length}\\
            & 3D & (2+1)D & iTGM &  3D & (2+1)D & iTGM & Pool \\
    \midrule
    Top 1 &  2  &  6  & 16  & 5 &  7.2  & 7.2  & 6.0 \\
    Top 2 &  6  &  7  & 12  & 7.8  &  8.1  & 8.6 &  5.7  \\
    Top 3 &  2  &  6  & 15  & 6  &  7.8  & 8.5 & 6.2  \\
    \bottomrule
  \end{tabular}
\end{table}

\noindent\textbf{Effectiveness of iTGM models}. 
In Table \ref{tab:model-stats}, we show the layer statistics for the best models. In the EvaNet architecture, iTGM layers have the longest average length (8.6). Further, our models have quite large temporal resolution of ∼368 frames on average (compared to I3D/S3D with 99 frames).
To further confirm the usefulness of the iTGM layer, we conduct several experiments. In Table \ref{tab:tgm-exps}, we show the results using iTGM layers with various temporal durations. Since we can increase the temporal length without changing the number of parameters, we can improve performance by simply taking longer temporal durations. We also compare to replacing all iTGM layers with (2+1)D layers and performing the architecture search without the iTGM layer as an option. Both restrictions degrade performance, confirming that iTGMs are needed. We also note that iTGM layers are most common in the best models (Table \ref{tab:model-stats}), further confirming their importance.

\begin{figure}
    \centering
    \includegraphics[width=0.83\linewidth]{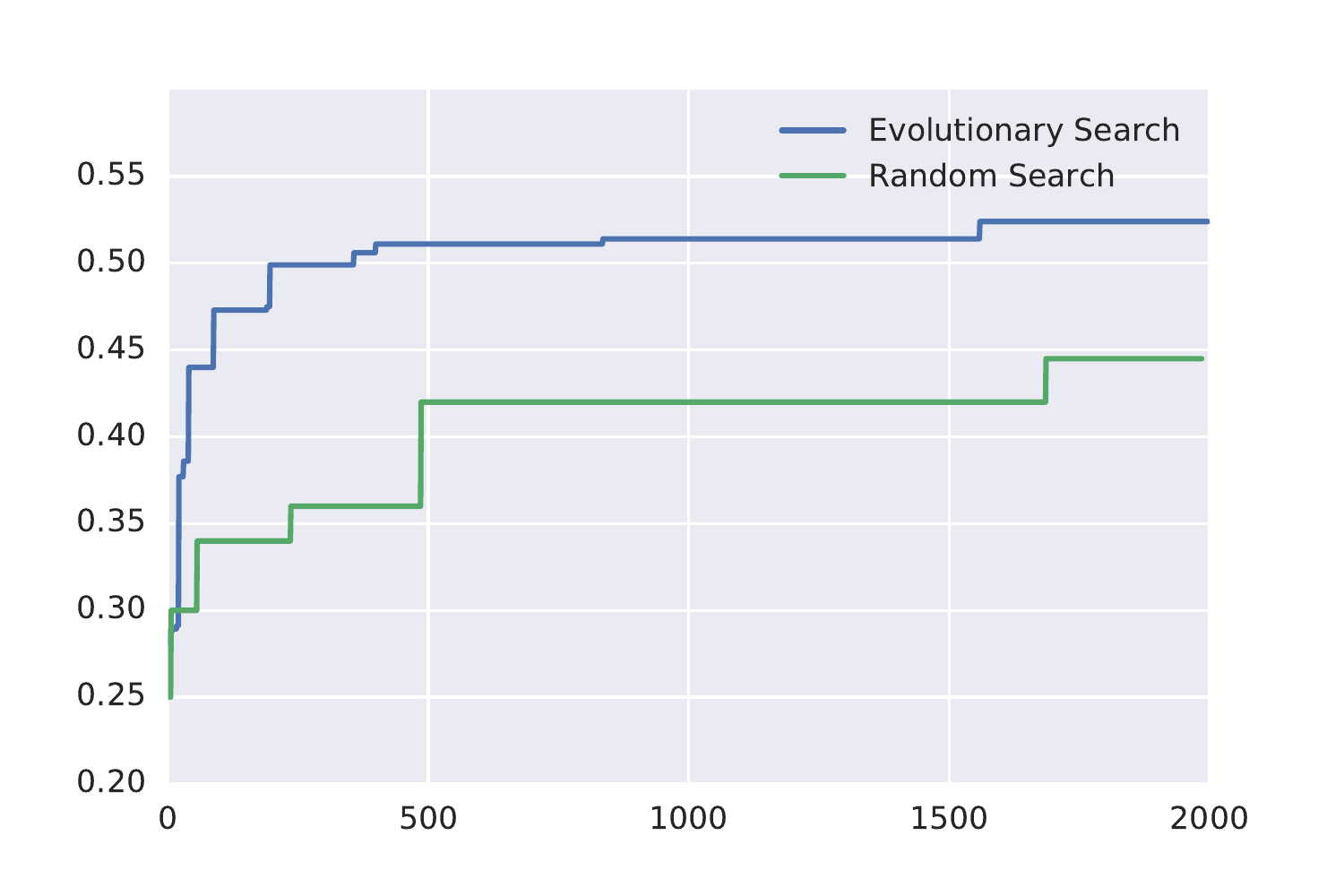}
    \caption{Random search vs. evolutionary algorithm on HMDB. X axis is number of rounds, Y axis is accuracy.}
    \label{fig:evolution-vs-random}
\end{figure}

\begin{table}
\small
  \caption{Experiments evaluating effect of iTGM layer on Kinetics.}
  \label{tab:tgm-exps}
  \centering
  \begin{tabular}{lc}
    \toprule
        Model    &  Accuracy \\
    \midrule
    iTGM ($L=3$) & 74.4 \\
    iTGM ($L=11$) & 74.9 \\
    EvaNet replacing iTGM with (2+1)D & 76.6 \\
    Arch Search without iTGM in space & 76.8 \\
    EvaNet & \textbf{77.2} \\
    \bottomrule
  \end{tabular}
\end{table}


\noindent\textbf{`Stretching' of iTGM layer}
Since the number of parameters of the iTGM layer is independent of length, we use a model from the Kinetics dataset and `stretch' the iTGM layers and apply it to Charades, which has activities with much longer temporal duration.
In Table \ref{tab:strech-exps}, we show the results using models with $L=3$ on Kinetics and stretched to $L=11$ on Charades, which shows similar performance.

\begin{table}
\small
  \caption{Stretching iTGM kernels from Kinetics to Charades.}
  \label{tab:strech-exps}
  \centering
  \begin{tabular}{lc}
    \toprule
        Model    &  mAP \\
    \midrule
    iTGM Baseline ($L=3$) & 33.8 \\
    iTGM Stretched ($L=11$) & 34.2 \\
    Kinetics EvaNet & 37.7 \\
    Kinetics EvaNet Stretched ($L=11$) & \textbf{38.1} \\
    Charades EvaNet & \textbf{38.1} \\
    \bottomrule
  \end{tabular}
\end{table}


\noindent\textbf{Evolution vs. random search.}
We compared our architecture evolution with random architecture search (Figure \ref{fig:evolution-vs-random}). We observe that both the evolution and the random search accuracies improve as they explore more samples (benefiting from the search space designed). However, the architecture evolution obtains much higher accuracy and much more quickly with few initial rounds of evolution, suggesting the mutations are being effective. 

\vspace{-0.2cm}
\section{Conclusion}
We present a novel evolutionary algorithm that automatically constructs architectures of layers exploring space-time interactions for videos. 
The discovered architectures are accurate, diverse and very efficient. 
Ensembling such models leads to further accuracy gains and yields faster and more accurate solutions than previous state-of-the-art models.
Evolved models can be used across datasets and to build more powerful models for video understanding. 

{\small
\bibliographystyle{ieee_fullname}
\bibliography{bib}
}

\clearpage
\newpage
\appendix

\section{Inception-like meta-architecture}

Similar to Fig.~4 of the main paper describing ResNet-like meta-architecture, we also describe the Inception-like meta-architecture we used. Fig.~\ref{fig:inception-meta} describes the architecture with nine modifiable modules, it has a `stem' of five modules. Note that the modules are evolved to be heterogeneous, having different combination of multiple conv. and pooling layers for each module. We describe multiple examples of such evolved architectures following the Inception-like meta-architectures in Figs.~\ref{fig:example-architecture-kinetics-1}, \ref{fig:example-architecture-kinetics-2}, \ref{fig:example-architecture-kinetics-3}, \ref{fig:example-architecture-kinetics-flow-1},
\ref{fig:example-architecture-kinetics-flow-2},
\ref{fig:example-architecture-kinetics-flow-3},
\ref{fig:example-architecture-charades-1},
\ref{fig:example-architecture-charades-2},
and \ref{fig:example-architecture-charades-3}.

\section{Evolution training details}

Our architecture evolution was done with 50 parallel workers. Each worker selects $S=25$ random samples from the population to generate one new child architecture based on the individual with the highest fitness (i.e., the parent). The architecture is trained using 12 GPUs on the training data. As training video CNNs is computationally expensive, during the search, we train the models with video segments of size $32\times 176\times 176$ (for HMDB and Kinetics) or $64\times 176\times 176$ (for Charades) where 32 and 64 are the number of frames. We use the batch size of 144 (12 per GPU).
Each newly generated child architecture is trained for 1000 iterations (i.e., it looks at 144000 samples), then evaluated with a separate validation set of 1000 examples. The classification accuracy measured using the validation becomes the `fitness' used in our algorithm.
We observed that relative recognition performances of the models (on the validation set) is stable after training for 1000 iterations, and we used this setting in our evolutionary algorithm to reduce the model training time necessary for the architecture evaluation.

\begin{figure}
    \centering
    \includegraphics[width=\linewidth]{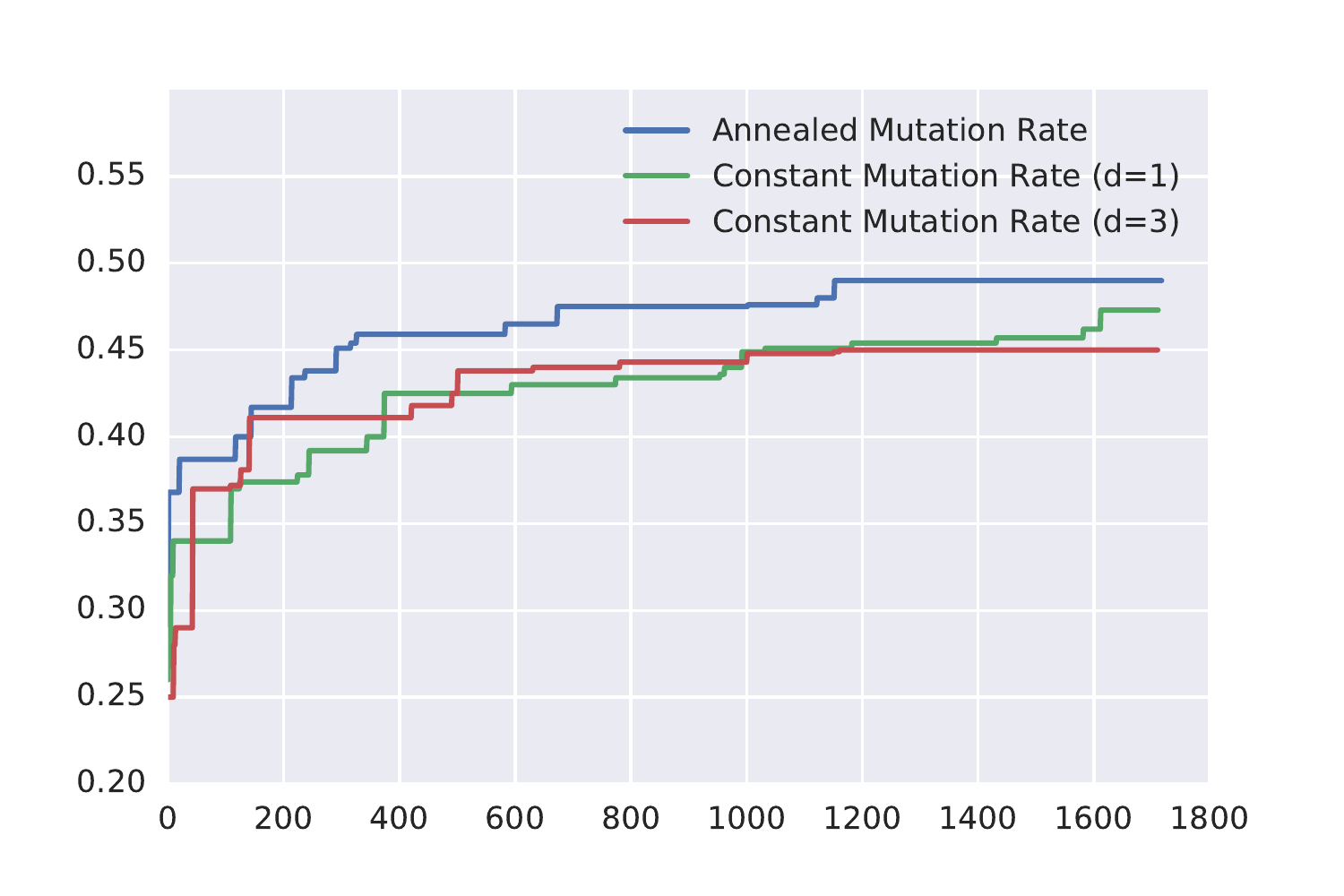}
    \caption{Comparison of the architecture search with various mutation rates. We observe that the constant rate takes longer to reach higher performance while the higher mutation rate initially learns faster, but plateaus at a lower value. Annealing the mutation rate based on the number of architectures evaluated provides the best performance. The $x$-axis is the number of evolutionary rounds and the $y$-axis is the accuracy after training for 1000 iterations.}
    \label{fig:mutation-rates}
\end{figure}

\section{Mutation rate}
In Fig. \ref{fig:mutation-rates}, we compare the architecture evolution done with a constant mutation rate of 1 or 3 (per round) and our annealed mutation rate. As we described in the main section of the paper, our evolutionary algorithm applies a set of random mutation operators at each round. In our annealed mutation rate strategy, the number of the mutation operators to apply is decided based on the evolution round $i$: it starts with $d = 7$ mutations initially and it is linearly decreased by $\lfloor i/r \rfloor$ where $r$ is 100 in our experimental setting. That is, at the $i$th round, a total of $max(\lceil d - i/r \rceil, 1)$ random mutations were applied to the parent. We find that the annealed mutation rate performs the best. Our strategy allows the search to explore more diverse architectures based on the best initial models, but then refine the top performing models after many evolution rounds.

\begin{figure*}
    \centering
    \includegraphics[width=\linewidth]{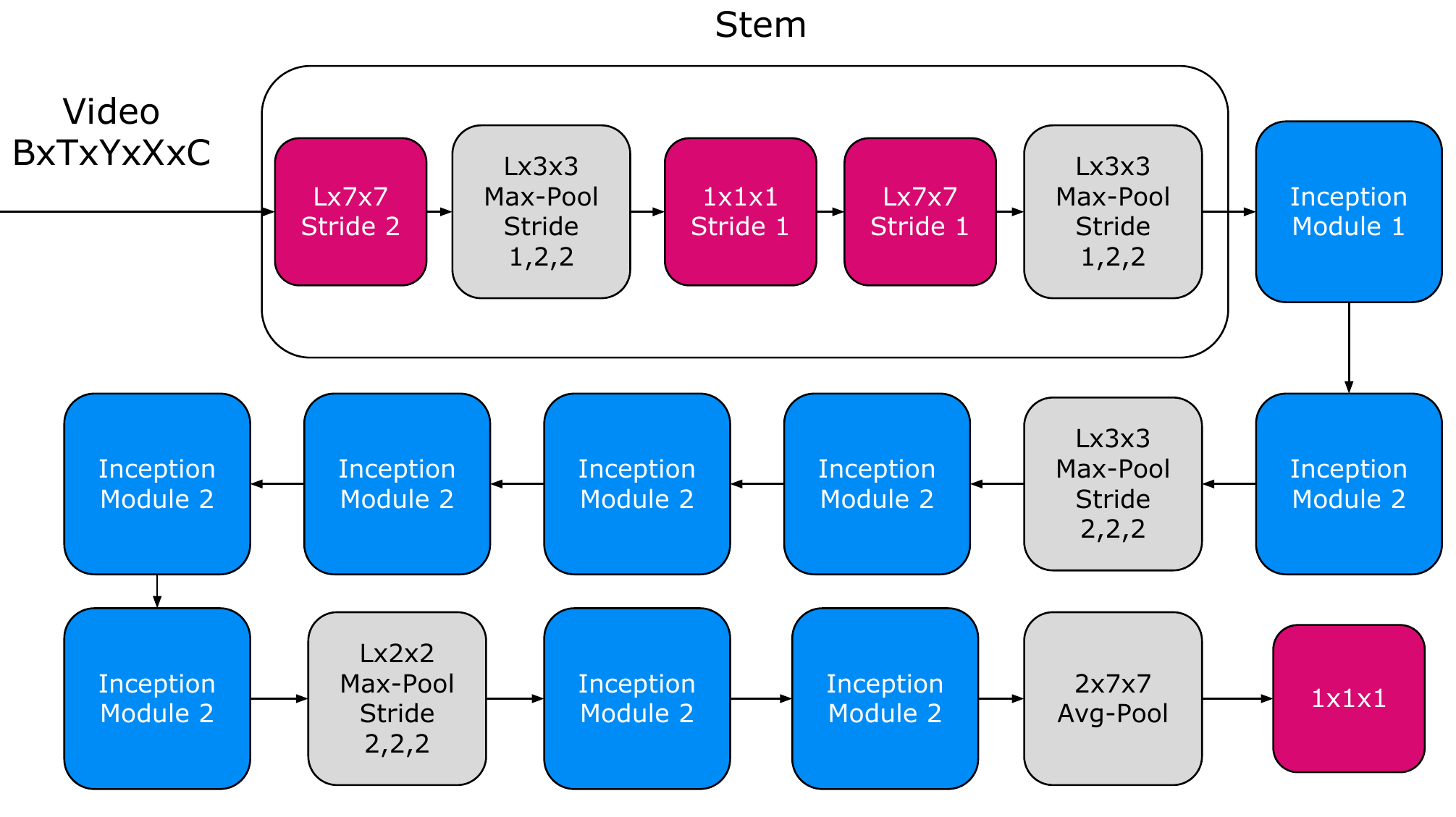}
    \caption{Inception-like meta-architecture.}
    \label{fig:inception-meta}
\end{figure*}

\section{Supplementary results and experiments}

In addition to the experimental results in the main paper, we below provide additional tables comparing our EvaNet against more detailed baselines sharing the same meta-architecture as our EvaNet. Table~\ref{tab:kinetics-simple-mutations} compares baseline and EvaNet models only using RGB input, optical flow input, and both. Table~\ref{tab:kinetics-sota} illustrates the performances of our EvaNet as well as previous works on two different settings of Kinetics-400. Note that Kinetics is periodically removing some of its training/validation/testing videos, and the accuracies of the approach thus changes depending on which version they were trained/tested on, thus the results are not directly comparable to the ones published on the larger set. Table~\ref{tab:charades-full} compares EvaNet with various baselines similar to Table~\ref{tab:kinetics-simple-mutations}, this time using the Charades dataset.

\begin{table}
  \caption{Kinetics performance comparison to baselines.}
  \label{tab:kinetics-simple-mutations}
  \centering
  \begin{tabular}{lccc}
    \toprule
            &  RGB & Flow & RGB+F \\
    \midrule
    \multicolumn{4}{l}{Baselines} \\
    3D Conv  & 70.6  & 62.1 & 72.6 \\
    (2+1)D Conv & 71.1 & 62.5 & 74.3 \\
    iTGM Conv  & 71.2 & 62.8 & 74.4 \\
    \hline
    3D-Ensemble  & & & 74.6 \\
    iTGM-Ensemble  & & & 74.7 \\
    \midrule
    \multicolumn{4}{l}{Top individual models from evolution}\\
    Top 1 & 71.9 & 63.8 & \textbf{76.4} \\
    Top 2 & 71.7 & 64.9 & 75.5 \\
    Top 3 & 72.9 & 64.8 & 75.7 \\
    \hline
    EvaNet &  &  & \textbf{77.2}\\
    \bottomrule
  \end{tabular}
\end{table}

\begin{table}
\caption{Kinetics-400 accuracy. Note that * are the reported numbers on the initial Kinetics dataset, which is no longer available. We report the numbers based on the new Kinetics version from Nov 2018. The new version has 8\% less training/validation videos.}
\label{tab:kinetics-sota}
\centering
\begin{tabular}{lccc}
\toprule
Method & \multicolumn{2}{c}{Kinetics-400} \\
 & new & old \\
\midrule
Two-stream I3D \cite{carreira2017quo}  & 72.6 & 74.1$^*$  \\ 
Two-stream (2+1)D \cite{tran2018closer} & - & 75.4$^*$  \\
ResNet-50 (2+1)D & 72.1 & - \\
ResNet-101 (2+1)D & 72.8 & -  \\
Two-stream S3D-G \cite{xie2017rethinking}  & 76.2 & 77.2$^*$\\
Non-local NN \cite{wang2017nonlocal} & - & 77.7$^*$\\
ResNet-50 + Non-local & 73.5 & - \\
\midrule
EvaNet (ours) & \textbf{77.2} & - \\
\bottomrule
\end{tabular}
\end{table}

\begin{table}
  \caption{Charades performance comparison to baselines, all initialized with ImageNet or Kinetics weights.}
  \label{tab:charades-full}
  \centering
  \begin{tabular}{lcc}
    \toprule
            &  ImageNet & Kinetics \\
    \midrule
    \multicolumn{3}{l}{Baselines} \\
    3D Conv  & 17.2  & 34.6 \\
    (2+1)D Conv & 17.1 & 34.7 \\
    iTGM Conv  & 17.2 & 34.9  \\
    \hline
    3D-Ensemble  & 17.4 & 35.2 \\
    iTGM-Ensemble  & 17.8 & 35.7 \\
    \midrule
    \multicolumn{3}{l}{Top individual models from evolution}\\
    Top 1 & 22.3 & \textbf{37.3} \\
    Top 2 & \textbf{24.1} & 36.8 \\
    Top 3 & 23.2 & 36.6  \\
    \hline
    EvaNet & \textbf{26.6} & \textbf{38.1}\\
    \bottomrule
  \end{tabular}
\end{table}

\section{Discovered architectures}
We here present diverse architectures evolved in the following figures. The color of each layer corresponds to a specific layer type. Check Fig.~1 of the main paper for the illustration. 

\begin{figure*}
    \centering
    \includegraphics[width=\linewidth]{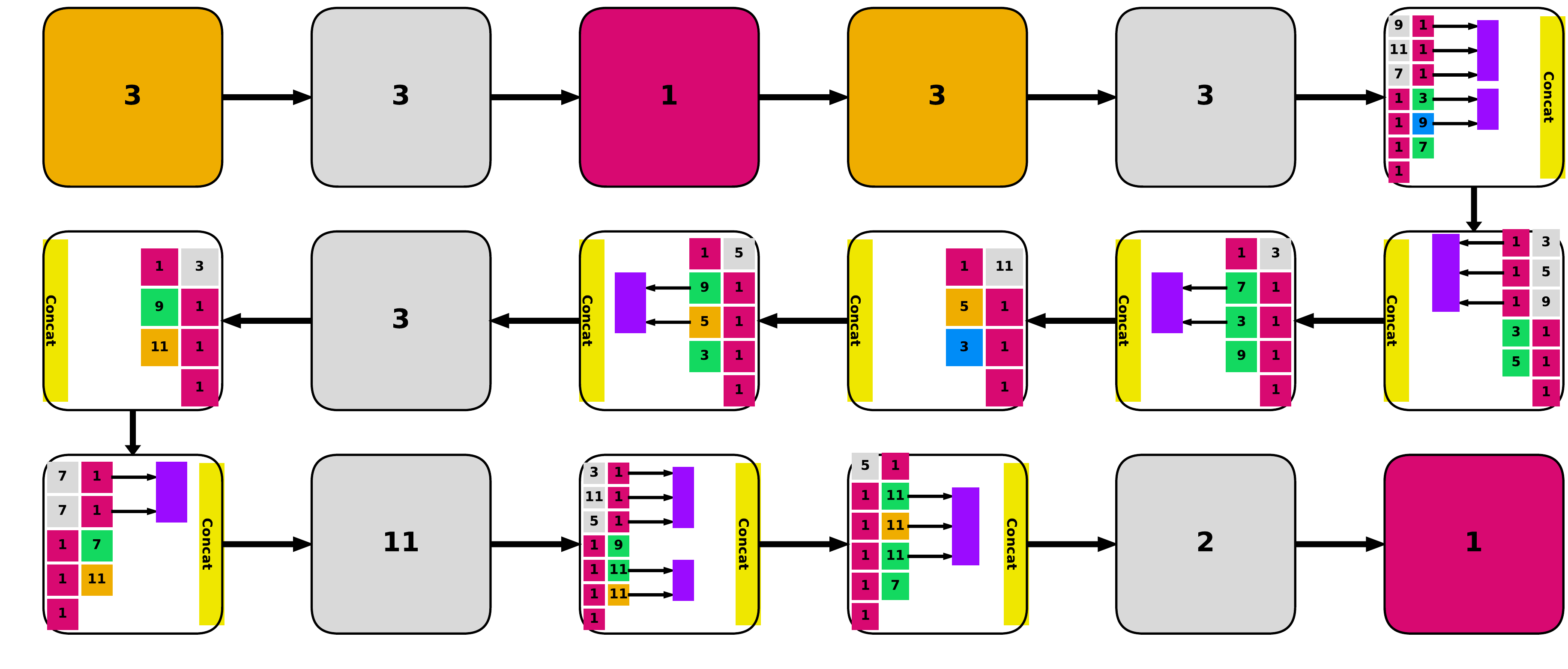}
    \caption{Kinetics RGB Top 1 with Inception Meta-architecture.}
    \label{fig:example-architecture-kinetics-1}
\end{figure*}

\begin{figure*}
    \centering
    \includegraphics[width=\linewidth]{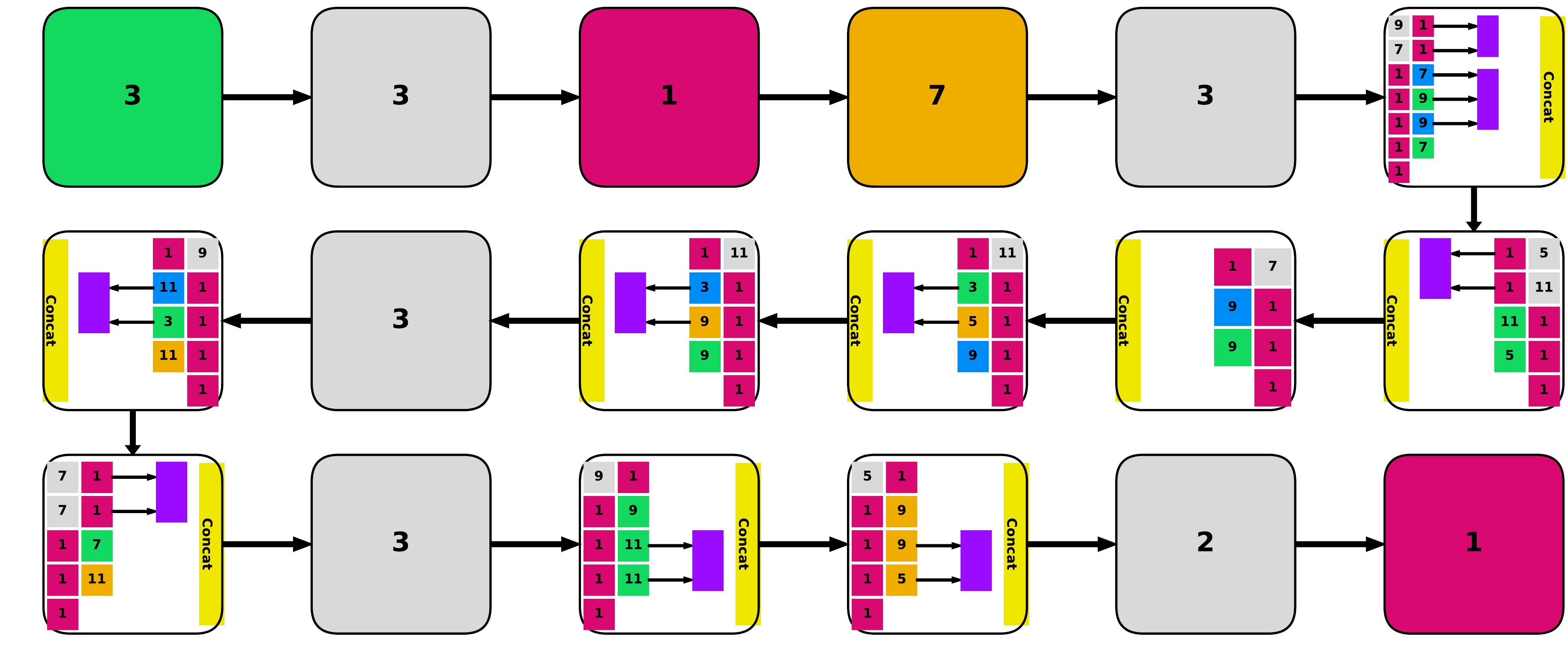}
    \caption{Kinetics RGB Top 2 with Inception Meta-architecture.}
    \label{fig:example-architecture-kinetics-2}
\end{figure*}

\begin{figure*}
    \centering
    \includegraphics[width=\linewidth]{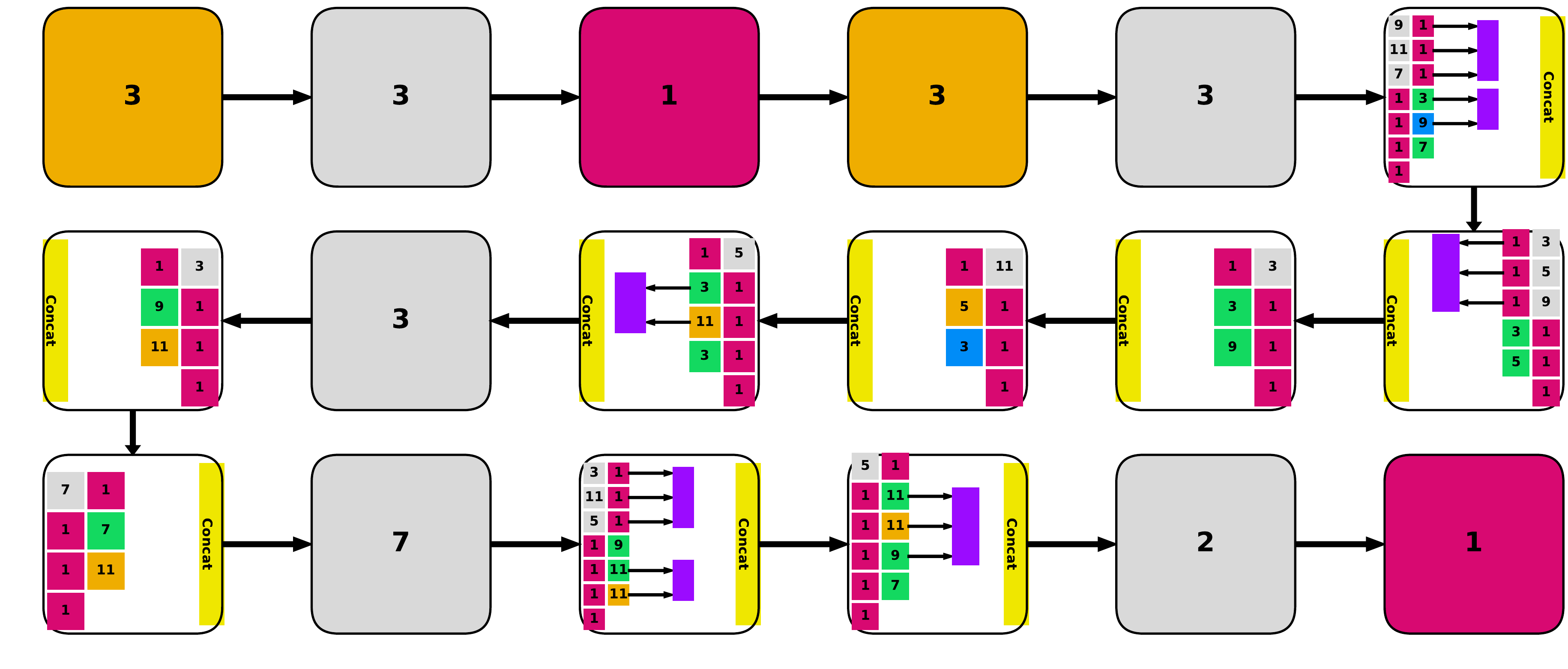}
    \caption{Kinetics RGB Top 3 with Inception Meta-architecture.}
    \label{fig:example-architecture-kinetics-3}
\end{figure*}

\begin{figure*}
    \centering
    \includegraphics[width=\linewidth]{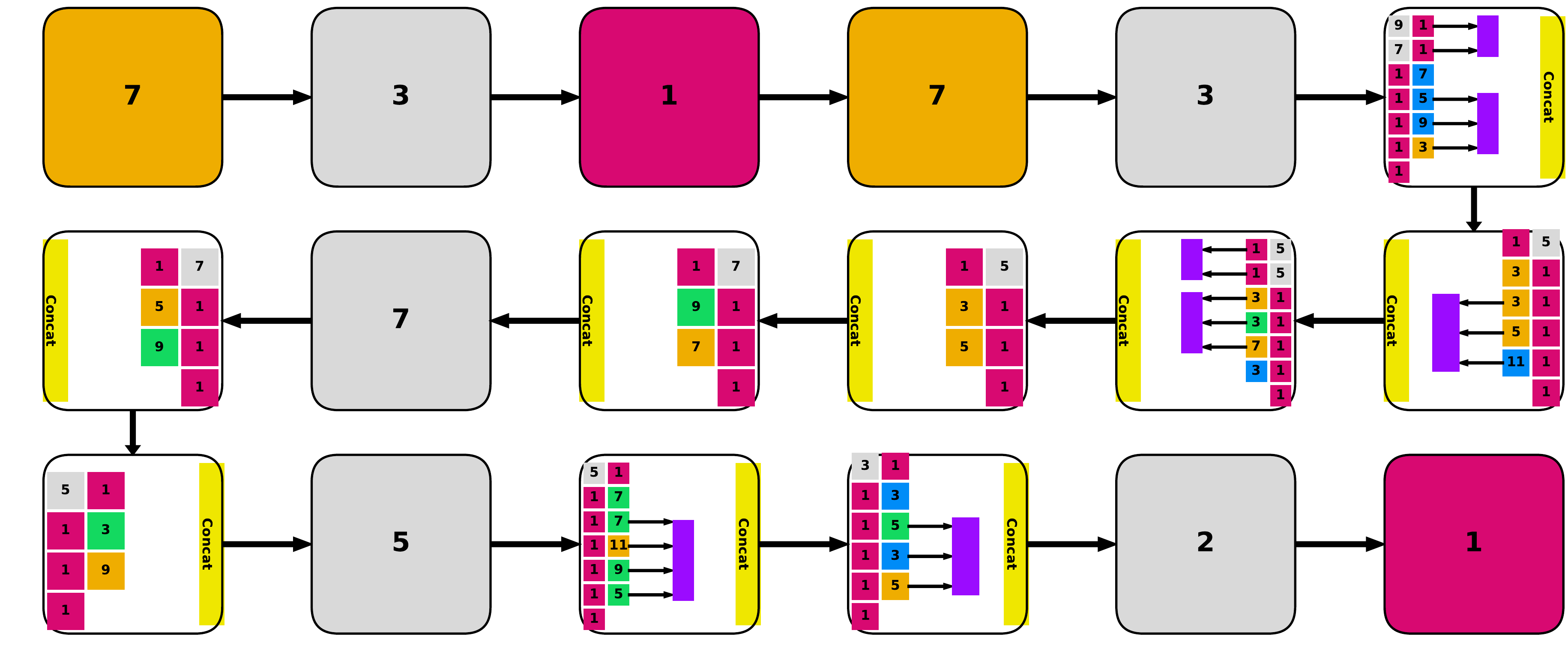}
    \caption{Kinetics optical flow Top 1 with Inception Meta-architecture.}
    \label{fig:example-architecture-kinetics-flow-1}
\end{figure*}

\begin{figure*}
    \centering
    \includegraphics[width=\linewidth]{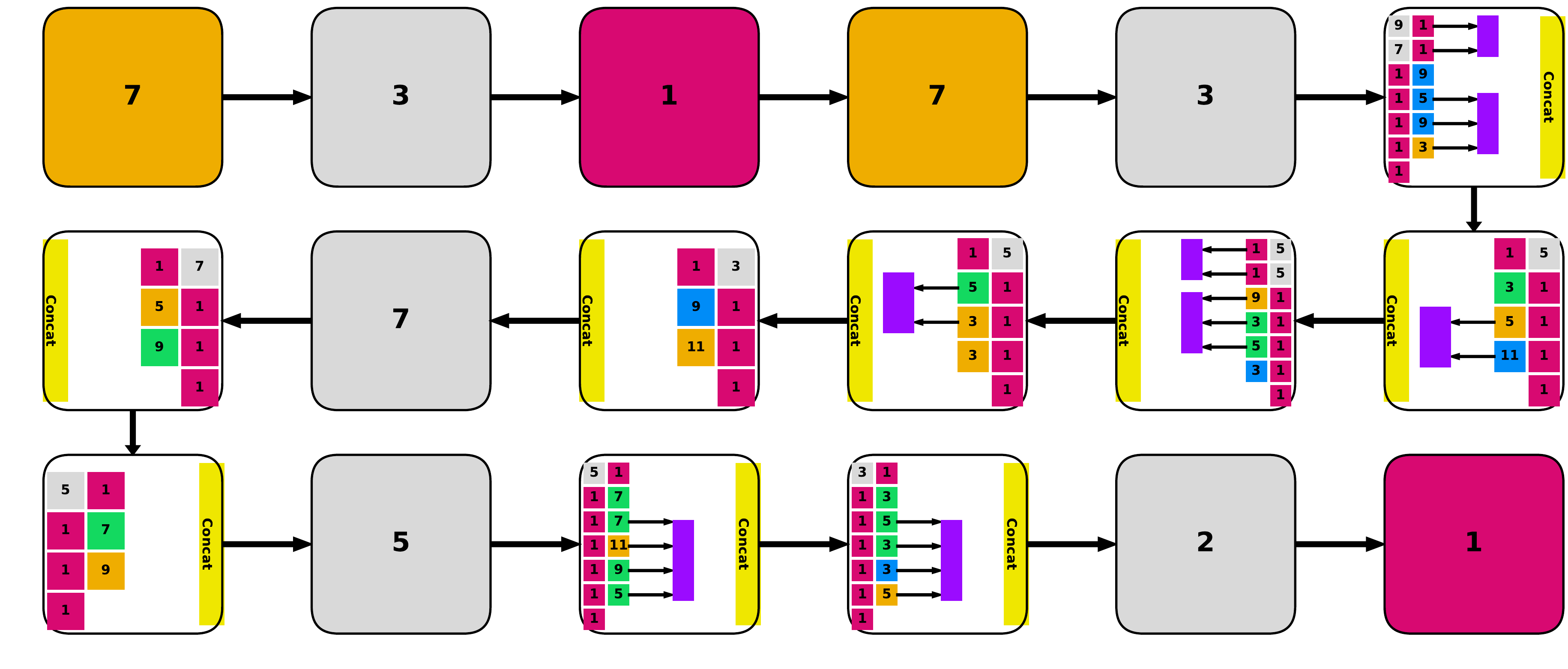}
    \caption{Kinetics optical flow Top 2 with Inception Meta-architecture.}
    \label{fig:example-architecture-kinetics-flow-2}
\end{figure*}

\begin{figure*}
    \centering
    \includegraphics[width=\linewidth]{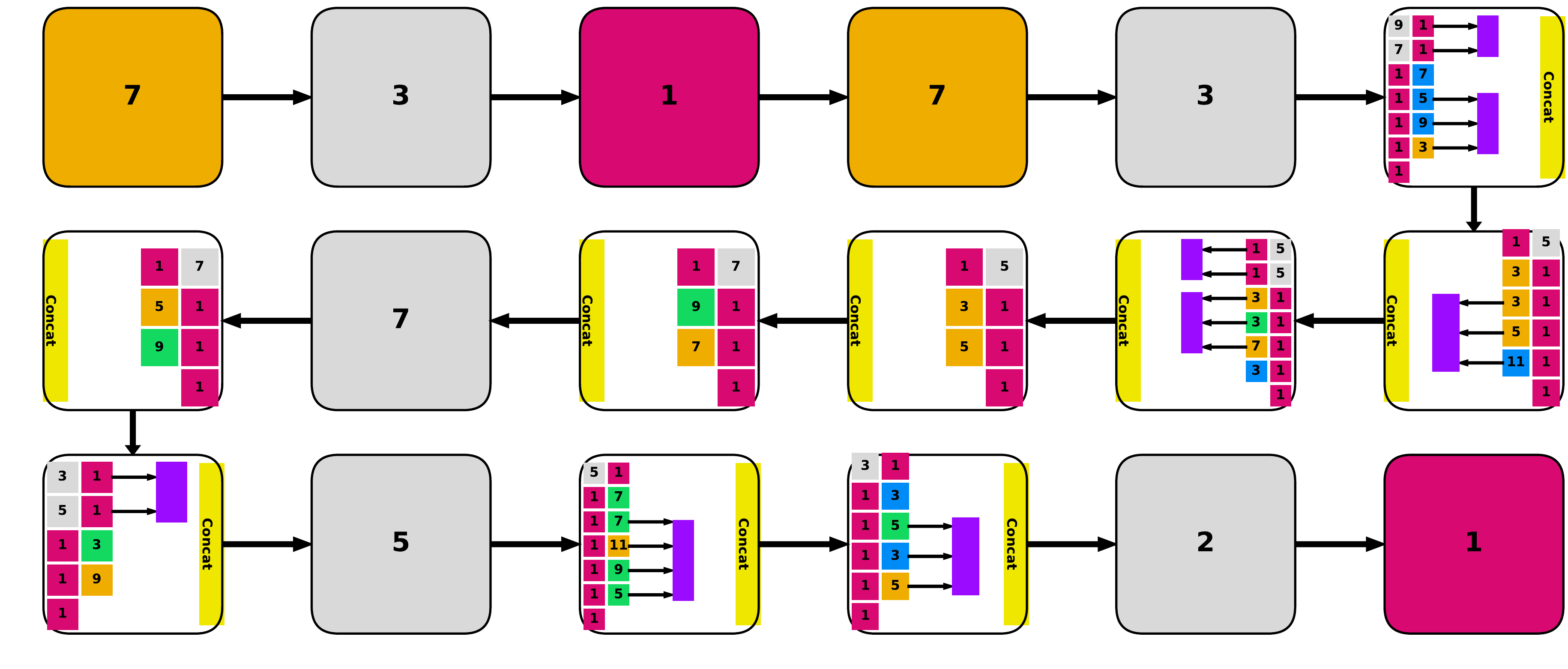}
    \caption{Kinetics optical flow Top 3 with Inception Meta-architecture.}
    \label{fig:example-architecture-kinetics-flow-3}
\end{figure*}

In Figures \ref{fig:example-architecture-kinetics-1}, \ref{fig:example-architecture-kinetics-2}, and \ref{fig:example-architecture-kinetics-3}, we show the Inception-like architectures found when searching on Kinetics using RGB inputs. We observe that the networks learn quite different architectures. For example, the third inception module is quite different in all three networks. In Figures \ref{fig:example-architecture-kinetics-flow-1}, \ref{fig:example-architecture-kinetics-flow-2}, and \ref{fig:example-architecture-kinetics-flow-3}, we illustrate the models found when searching on Kinetics using optical flow as input. When using optical flow as input, we observe that the architectures perfer to use layers with shorter temporal durations, using very few layers with size 11 and 9 when compared to the RGB networks.  (2+1)D conv layers and iTGM layers were used much more commonly in both RGB and optical flow architectures. Parallel space-time conv and pooling layers with different temporal lengths were also very commonly observed.

\begin{figure*}
    \centering
    \includegraphics[width=\linewidth]{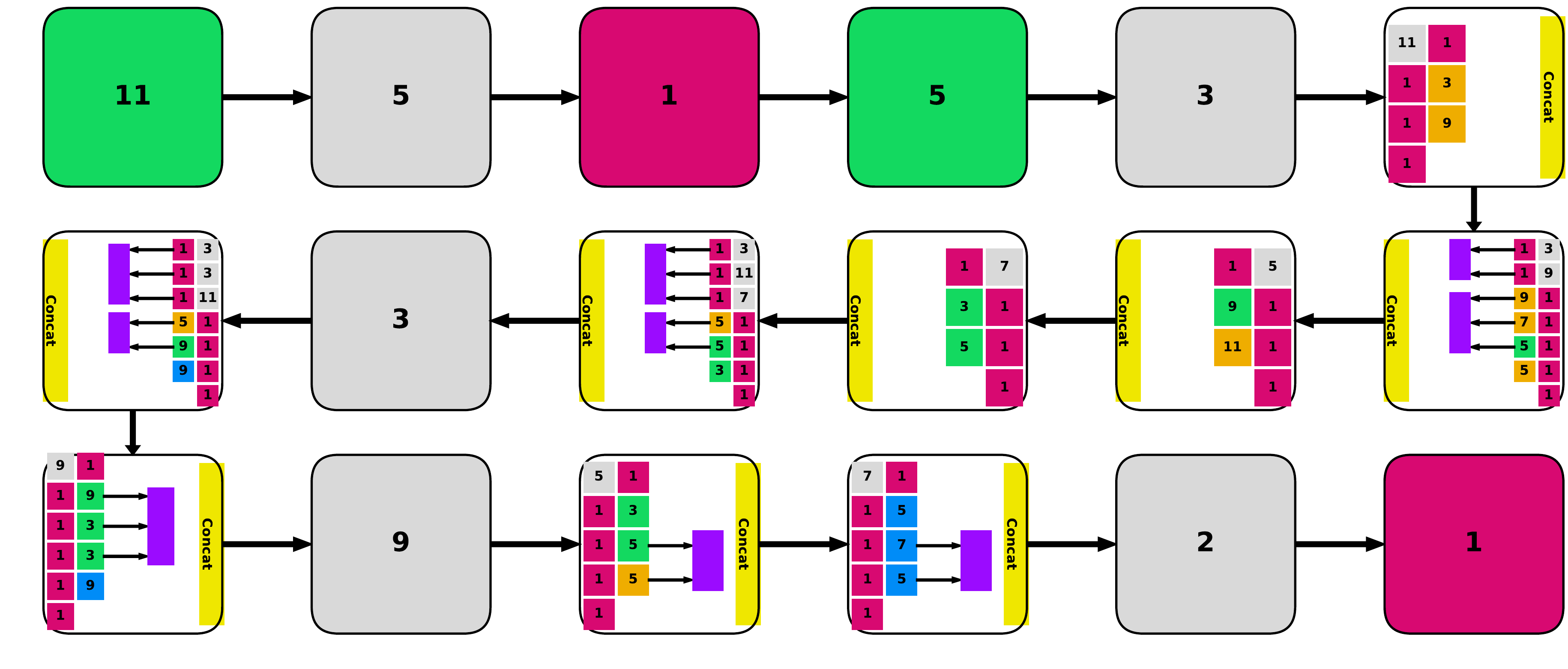}
    \caption{Charades RGB Top 1.}
    \label{fig:example-architecture-charades-1}
\end{figure*}

\begin{figure*}
    \centering
    \includegraphics[width=\linewidth]{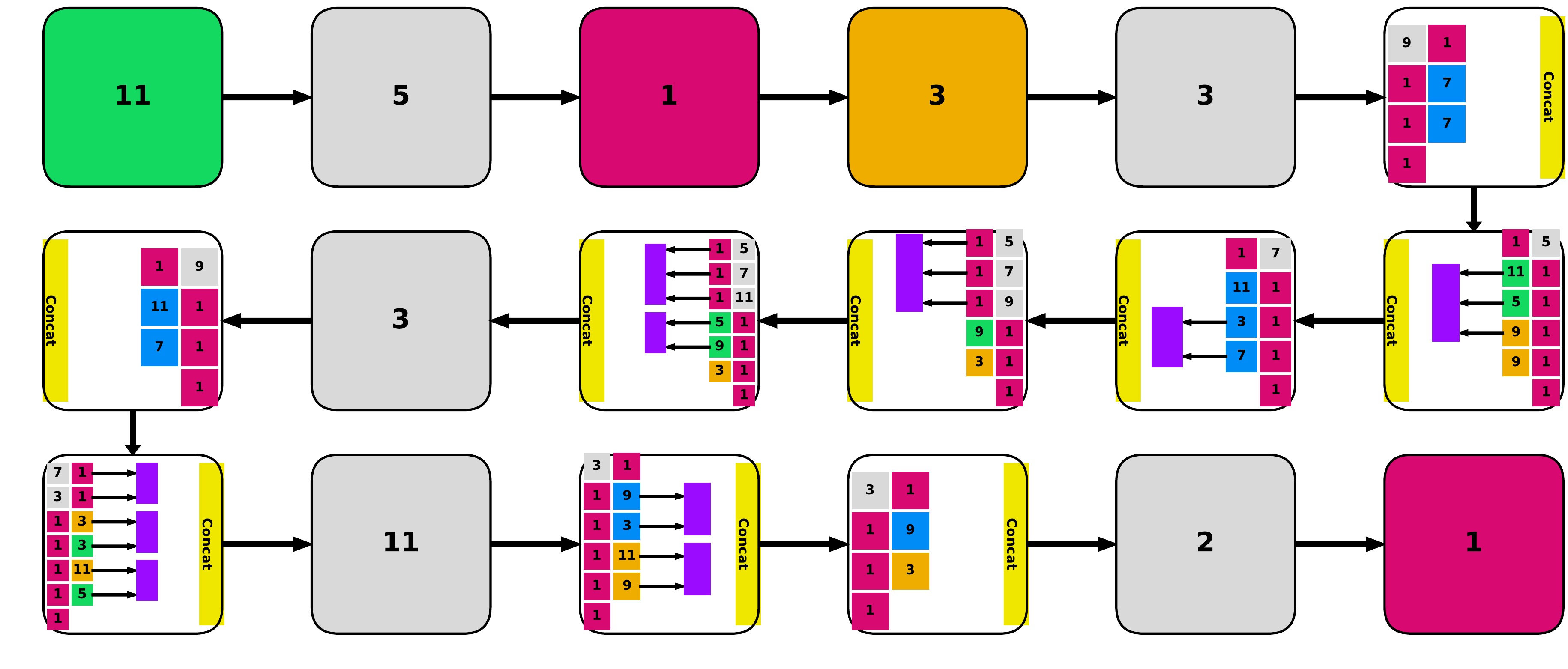}
    \caption{Charades RGB Top 2.}
    \label{fig:example-architecture-charades-2}
\end{figure*}

\begin{figure*}
    \centering
    \includegraphics[width=\linewidth]{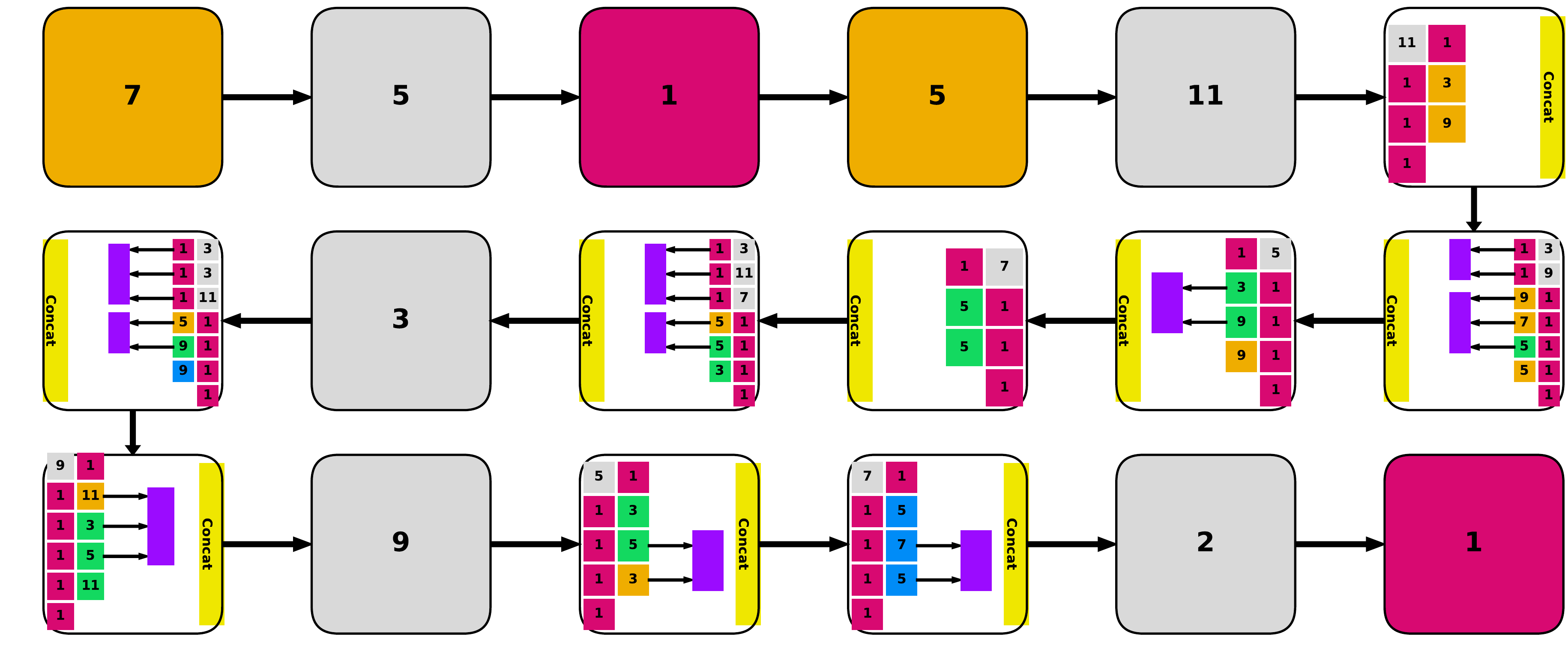}
    \caption{Charades RGB Top 3.}
    \label{fig:example-architecture-charades-3}
\end{figure*}

In Figures \ref{fig:example-architecture-charades-1}, \ref{fig:example-architecture-charades-2}, and \ref{fig:example-architecture-charades-3}, we illustrate the Inception-like architectures evolved on Charades. We observe that on Charades, the architectures generally capture longer temporal intervals (e.g., the first layer has size 11) and many layers contain longer kernels (i.e., 9 and 11) especially compared to the architectures found on Kinetics. 

\begin{figure*}
    \centering
    \includegraphics[width=\linewidth]{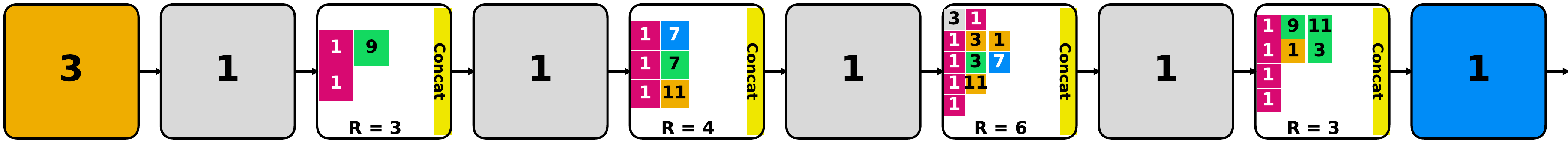}
    \caption{Kinetics RGB Top 1 with ResNet Meta-Architecture.}
    \label{fig:example-architecture-kinetics-1-res}
\end{figure*}

\begin{figure*}
    \centering
    \includegraphics[width=\linewidth]{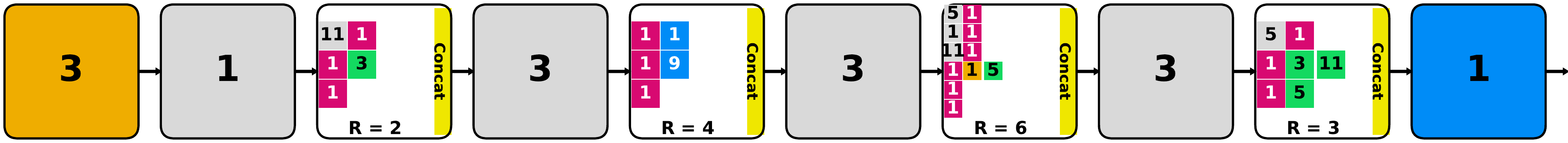}
    \caption{Kinetics RGB Top 2 with ResNet Meta-Architecture.}
    \label{fig:example-architecture-kinetics-2-res}
\end{figure*}

\begin{figure*}
    \centering
    \includegraphics[width=\linewidth]{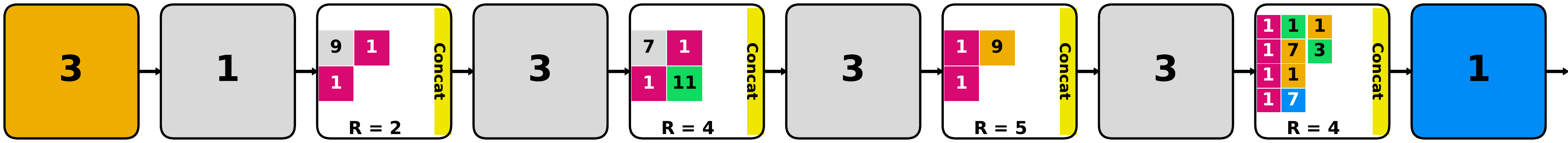}
    \caption{Kinetics RGB Top 3 with ResNet Meta-Architecture.}
    \label{fig:example-architecture-kinetics-3-res}
\end{figure*}

\begin{figure*}
    \centering
    \includegraphics[width=\linewidth]{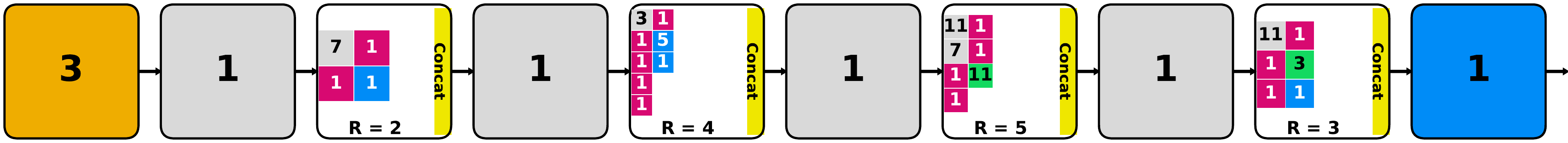}
    \caption{Kinetics optical flow Top 1 with ResNet Meta-Architecture.}
    \label{fig:example-architecture-kinetics-flow-1-res}
\end{figure*}

\begin{figure*}
    \centering
    \includegraphics[width=\linewidth]{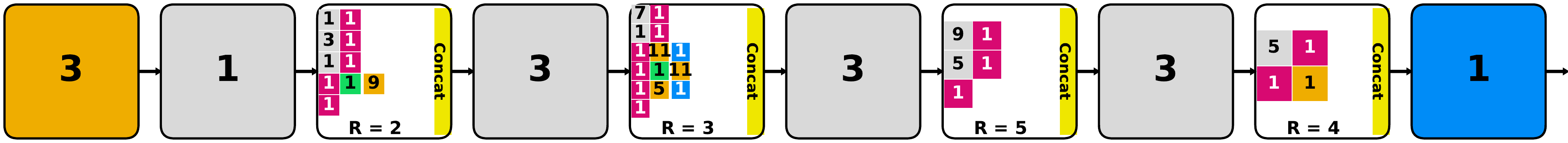}
    \caption{Kinetics optical flow Top 2 with ResNet Meta-Architecture.}
    \label{fig:example-architecture-kinetics-flow-2-res}
\end{figure*}

\begin{figure*}
    \centering
    \includegraphics[width=\linewidth]{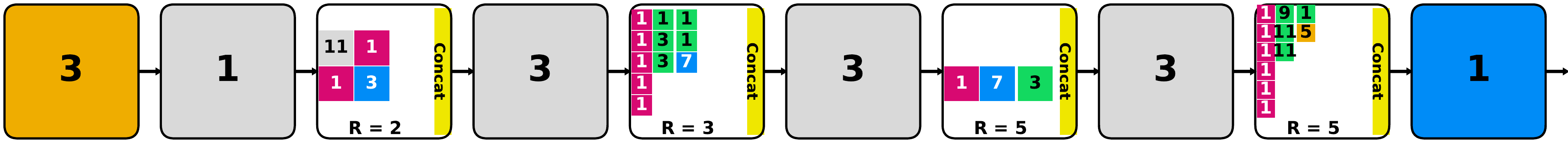}
    \caption{Kinetics optical flow Top 3 with ResNet Meta-Architecture.}
    \label{fig:example-architecture-kinetics-flow-3-res}
\end{figure*}

In addition, we also illustrate evolved architectures based on our ResNet-like meta-architecture. Similar to the above mentioned figures, we show three examples of evolved architectures per Kinetics-RGB and Kinetics-Flow. Figs.~\ref{fig:example-architecture-kinetics-1-res}, Figs.~\ref{fig:example-architecture-kinetics-2-res}, Figs.~\ref{fig:example-architecture-kinetics-3-res}, 
Figs.~\ref{fig:example-architecture-kinetics-flow-1-res}, 
Figs.~\ref{fig:example-architecture-kinetics-flow-2-res}, 
and Figs.~\ref{fig:example-architecture-kinetics-flow-3-res} show the architectures.


{\small
\bibliographystyle{ieee}
\bibliography{bib}
}
\end{document}